\documentclass[letterpaper, 10 pt, conference, twoside]{ieeetran}

\IEEEoverridecommandlockouts

\usepackage[bookmarks=true]{hyperref}

\usepackage{xcolor}
\usepackage{times}
\usepackage{epsfig}
\usepackage{graphicx}
\usepackage{tabularx,ragged2e,booktabs}

\usepackage{amsmath}
\usepackage{amssymb}
\usepackage{bbm}
\usepackage[misc]{ifsym}
\usepackage[export]{adjustbox}
\usepackage{float}
\usepackage{wrapfig}
\usepackage{lscape}
\usepackage{rotating}
\usepackage{epstopdf}

\usepackage{verbatim}
\usepackage{booktabs}
\usepackage{multirow}
\usepackage{makecell}
\usepackage{authblk}
\usepackage{hyphenat} 

\usepackage{pgf}

\DeclareMathOperator*{\argmax}{arg\,max}

\usepackage{graphics}
\usepackage{pifont}

\usepackage{times}
\usepackage{soul}
\usepackage{url}
\usepackage{graphicx}
\usepackage{amsmath}
\usepackage{booktabs}
\usepackage{balance} 
\usepackage[T1]{fontenc}    
\usepackage{wrapfig}
\usepackage{fancybox}
\usepackage{tikz}
\usepackage{algpseudocode}
\usepackage{algorithmicx}
\usepackage[ruled,vlined,linesnumbered,noend]{algorithm2e}
\usepackage{diagbox}
\usepackage{multirow}
\usepackage{amsthm}
\usepackage{amsfonts}       
\usepackage{nicefrac}       

\usepackage{graphbox}

\graphicspath{{figures/}}

\newcommand{\revision}{black}

\makeatletter
\def\ps@IEEEtitlepagestyle{%
  \def\@oddhead{\mycopyrightnotice}%
  \def\@oddfoot{\hbox{}\@IEEEheaderstyle\leftmark\hfil\thepage}\relax
  \def\@evenhead{\@IEEEheaderstyle\thepage\hfil\leftmark\hbox{}}\relax
  \def\@evenfoot{}%
}
\def\mycopyrightnotice{%
  \begin{minipage}{\textwidth}
  \centering \scriptsize
  This article has been accepted for publication in the IEEE Robotics and Automation Letters (RA-L) Journal. Copyright~\copyright~20XX IEEE.  Personal use of this material is permitted.  Permission from IEEE must be obtained for all other uses, in any current or future media, including reprinting/republishing this material for advertising or promotional purposes, creating new collective works, for resale or redistribution to servers or lists, or reuse of any copyrighted component of this work in other works.
  \end{minipage}
}
\makeatother

\begin{document}
\title{\LARGE {CQLite: Communication-Efficient Multi-Robot Exploration Using Coverage-biased Distributed Q-Learning}}

\author{Ehsan Latif \and Ramviyas Parasuraman
\thanks{The authors are with the Heterogeneous Robotics Research Lab, School of Computing, University of Georgia, Athens, GA 30602, USA.}
\thanks{Author emails: \tt{\{ehsan.latif;ramviyas\}}@uga.edu}
}

\maketitle              

\begin{abstract}
Frontier exploration and reinforcement learning have historically been used to solve the problem of enabling many mobile robots to autonomously and cooperatively explore complex surroundings. These methods need to keep an internal global map for navigation, but they do not take into consideration the high costs of communication and information sharing between robots. This study offers CQLite, a novel distributed Q-learning technique designed to minimize data communication overhead between robots while achieving rapid convergence and thorough coverage in multi-robot exploration. The proposed CQLite method uses ad hoc map merging, and selectively shares updated Q-values at recently identified frontiers to significantly reduce communication costs. The theoretical analysis of CQLite's convergence and efficiency, together with extensive numerical verification on simulated indoor maps utilizing several robots, demonstrates the method's novelty. With over 2x reductions in computation and communication alongside improved mapping performance, CQLite outperformed cutting-edge multi-robot exploration techniques like Rapidly Exploring Random Trees and Deep Reinforcement Learning. Related codes are open-sourced at \url{https://github.com/herolab-uga/cqlite}.

\end{abstract}


\section{Introduction}
\label{sec:intro}

Map-based coverage and exploration is a significant problem of interest in the robotics and multi-robot systems (MRS) community \cite{burgard2005coordinated}. In this problem, robots continuously explore to obtain the full environmental map in an unknown bounded environment. It can be helpful in various applications, including search and rescue, domestic service, survey and operations, field robotics, etc. Autonomous exploration and surveillance solutions can also demonstrate the adaptability of the MRS since robots can carry out these missions in different and uncharted areas. 
In such applications, the robots need efficient wireless network connectivity for robust cooperation in uncertain environments \cite{parasuraman2017new}.

Recent works have been influential in realizing an efficient exploration objective. For example, information-based methods (e.g., \cite{fang2019autonomous}) typically use the Shannon entropy to describe the uncertainty of the environmental map and construct the optimization problems such that the robot's control variable (e.g., velocity) is continuously optimized during the exploration process. On the other hand, frontier-based methods (e.g., \cite{dai2020fast}) involve deciding the robot's next move (or path) by searching the frontier points on the border of free and unknown points. Often, these methods only produce approximate solutions due to optimization.

Integrating learning with planning solutions is promising, especially for robot exploration \cite{latif2023communication,shrestha2019learned}. In the reinforcement learning (RL) paradigm, robots can continuously improve competence and adapt to the dynamics of natural surroundings by observing the results of navigational choices made in the actual world \cite{zhang2022multi}. On the other hand, cooperation among robots in an MRS can help achieve a complex mission through simple distributed approaches \cite{tolstaya2021multi,yang2019self}. 

\begin{figure}[t]
\centering
 \includegraphics[width=0.99\linewidth]{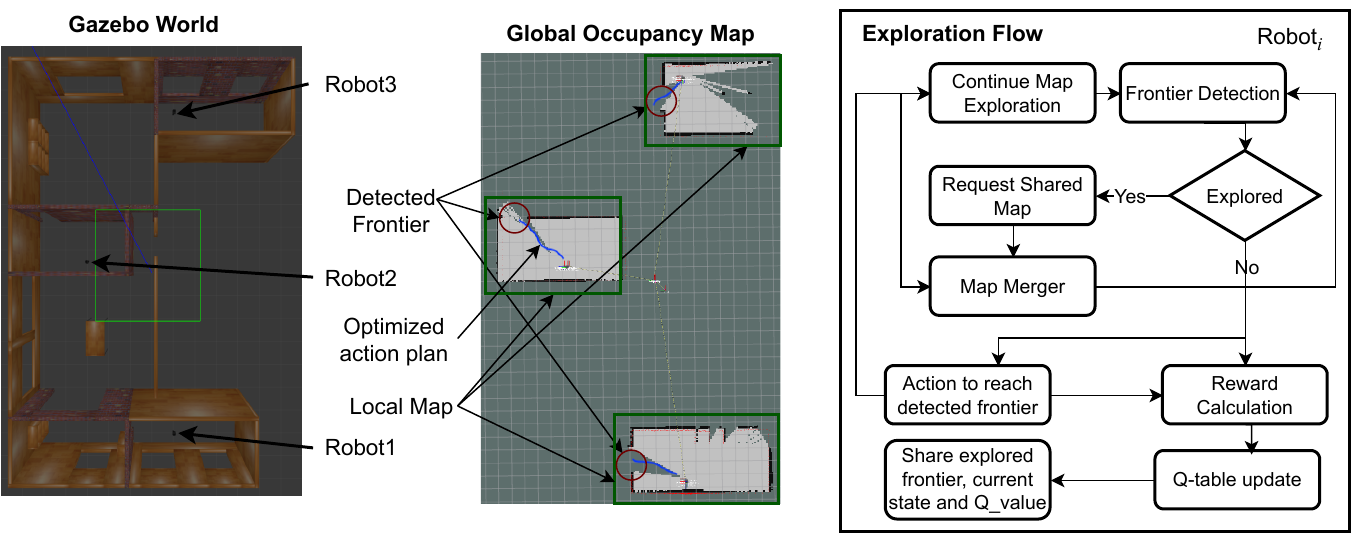}
 \caption{Overview of the distributed CQLite method for efficient multi-robot exploration, shown with an illustrative simulation.}
 \label{fig:overview}
\end{figure}

This paper explores the intersection between learning and cooperation, designs a combined solution to achieve efficient map exploration, and provides theoretical support for fast convergence and time complexity. We leverage the benefits of learning-based paradigms for joint exploration. We aim to create a distributed algorithm that gains knowledge through robot-robot information sharing while minimizing communication and computing overheads.
Specifically, we utilize a distributed Q-learning methodology with a coverage-biased reward function with a light communication and information fusion strategy. In our approach, we reduce communication complexity by sharing only the current state information, i.e., Q-value, instead of the complete Q-table as done in \cite{sadhu2018efficient} and explored frontier.
Fig.~\ref{fig:overview} provides an overview of the proposed method implemented in the Robot Operating System (ROS) framework.
The main contributions of this paper are summarized below.


\begin{itemize}
    \item We propose a novel distributed coverage-biased Q-learning approach (\textit{CQLite}) for efficient multi-robot map exploration, leading to a significant reduction of the communication and computation overheads at each robot. In our approach, we reduce communication complexity by sharing only the current state information, i.e., Q-value, instead of the complete Q-table as done in \cite{sadhu2018efficient} and apply map merging in an ad-hoc manner. 
    \item We substantiate the potential of our method with theoretical proofs of fast convergence in learning. 
    \item With extensive simulation experiments, we evaluate the performance of our approach against two state-of-the-art (SOTA) multi-robot exploration methods: Rapidly-exploring Random Trees (RRT) for Optimized Exploration \cite{zhang2020rapidly} and Deep Reinforcement Learning (DRL) for Voronoi-based Exploration \cite{hu2020voronoi}. 
    \item We open source\footnote{\url{https://github.com/herolab-uga/cqlite}} the CQLite as a ROS package for use and further development by the robotics community.
    \item We confirm the findings through \textbf{real-world robot experiments}. The video of sample simulation experiments and \textbf{real-world demonstrations} are available at \url{https://youtu.be/n3unL1nuieQ}.

\end{itemize}

The key idea behind the CQLite is that it uses a coverage-biased reward function to perform efficient exploration by sharing limited information among robots in a distributed fashion. Our method achieves fast convergence with the best coverage performance, reduced communication, and update costs compared to the baselines.

\section{Related Work}
\label{sec:relatedwork}

Map exploration problems focus on frontier-based and learning-based coverage planning approaches. A robot can be greedily pushed in an occupancy grip-map to the closest boundaries \cite{gao2018improved} or to the most uncertain (or informative) regions \cite{bouman2020autonomous}. 
In frontier-based strategies, the robots will look to expand coverage into the unexplored regions by choosing the next waypoints based on the frontier of the explored map boundaries \cite{latif2023seal}. For instance, in \cite{zhang2020rapidly}, the multi-robot map exploration objective is integrated into an optimization framework incorporating Rapidly-exploring Random Trees (RRTs) to increase the effectiveness and efficiency of multi-robot map exploration. However, the constraints of such frontier-based approaches are the computing expense of the optimization methods and the possibility of non-optimal outcomes resulting from RRTs' stochastic characteristics.

Researchers also presented communication-efficient solutions for exploration in multi-robot systems. For instance, Zhang et al. \cite{zhang2022mr} introduced the MR-TopoMap based on a topological map, which can independently explore the robot's surroundings while sporadically exchanging topological maps when communication is possible. However, path planning through topological mapping can lead to a sub-optimal path and, specifically, in the case where robots start exploring from the same position, exploring the same map, making it difficult to divide the map into topologies.
Masaba and Li \cite{masaba2021gvgexp} proposed an exploration algorithm using the topology of the generalized Voronoi graphs made efficient through a recurrent and lean communication strategy. Corah et al. \cite{corah2019communication} use information-based distributed planning considering communication restrictions. However, the planner's finite-horizon nature could lead to suboptimal exploration paths because it doesn't consider long-term planning beyond the given horizon, making it more difficult for the system to make decisions based on knowledge in the future. This might prevent robots from efficiently exploring or discovering key regions of interest.
More recently, Gao et al. \cite{gao2022meeting} reduced inter-robot communication costs by utilizing a mission-based protocol and centralized planning, where the former can actively disconnect robots to proceed with distributed (independent exploration) and the latter will help them achieve rendezvous to reconnect and share information. However, computing the super-frontier information is computationally expensive, and the active disconnection strategy may limit the robots from sharing other critical data during the mission.

A body of research concentrates on Reinforcement Learning (RL) and Q-learning for multi-robot tasks, modifying the learning mechanism in low communication scenarios for better navigation and exploration \cite{serra2020whom,latif2023communication} and utilizing deep reinforcement learning to achieve optimality in robotic exploration \cite{han2020cooperative}. 
A Deep RL (DRL) approach for cooperative multi-robot exploration using Voronoi cells was proposed in \cite{hu2020voronoi}. Despite its intriguing concept, it was constrained by training difficulties and sub-optimal solution tendencies.
In another study \cite{yu2019navigation}, researchers suggested using an improved Q-learning algorithm in RL map navigation to stop robots from lingering in the past. 
However, these methods call for frequent map merging, which raises the cost of updates. 
Further, DRLs have shown promise in some problem spaces, but they frequently offer less-than-ideal solutions outside those contexts. They cannot guarantee convergence in infinite horizons.

To address these gaps in the literature, CQLite considers an efficient information transfer mechanism combined with distributed Q-learning with a coverage-biased reward function for efficient multi-robot cooperation to solve map exploration tasks.
CQLite departs from RRT (frontier-based) and DRL (learning-based) regarding exploration strategy by reducing recurrent frontier exploration to avoid mapping overlap and Q-learning update strategy for communication efficiency by only sharing and utilizing recently calculated Q-value to the robots, respectively. Additionally, in both RRT and DRL, robots share locally explored maps on every iteration and apply map merging, which consequently gives rise to computational complexity. We reduce this overhead by only sharing and applying map merging in an ad-hoc manner.
By incorporating these novelties, our approach provides significant improvement in computation and communication efficiency, even in cases of limited connectivity scenarios.

It is worth noting that the objectives of cooperative simultaneous localization and mapping (SLAM) techniques and exploration approaches are fundamentally different. 
The SLAM problem focuses on accurately building and merging the map, while the exploration problem focuses on using the available map to determine waypoints to maximize coverage area. 
Cooperative SLAM techniques emphasize communication effectiveness. Although computational efficiency is still a problem, Liu et al. \cite{liu2022robust} presented a multi-agent SLAM technique that lowers bandwidth use. Others use spectral graph analysis for cooperative mapping but overlook the computational costs of graph formation and optimization \cite{bernreiter2022collaborative}. In contrast, others concentrate on lifelong localization and mapping but fail to optimize the communication and computational cost \cite{zhao2021general}. Cooperative RL techniques, as those in \cite{jia2021lvio}, have difficulty keeping up with the rising computational complexity of growing state spaces. 
In our work, we use an existing map merging method\footnote{\url{https://wiki.ros.org/multirobot_map_merge}} from the literature to perform multi-robot SLAM. At the same time, our proposed CQLite is designed to maximize exploration and lower communication and computation. Specifically, our CQLite method addresses the above limitations by delivering computational and communication efficiency through selective data sharing and utilizing effective Q-learning to determine the best exploration strategy.

\begin{figure*}[t]
    \centering
\includegraphics[width=0.98\linewidth]{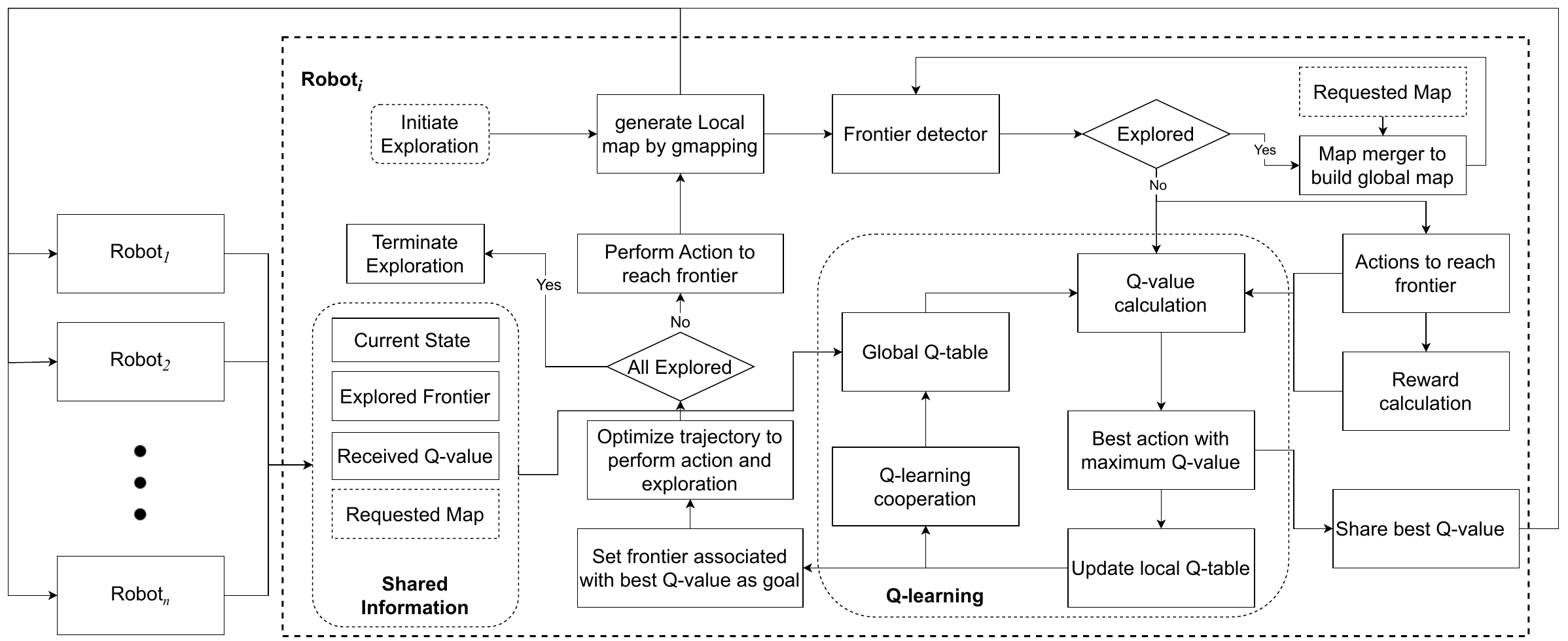}
\caption{System architecture of CQLite distributed across several robots. It shows the Robot $i$'s process showing the mapping, frontier detection, and Q-learning operations along with the communication of local map and updated Q-value information to $n$ connected robots.}
    \label{fig:system-architecture}
\end{figure*}

\section{Proposed Approach}
\label{sec:problem}
We consider $n$ connected robots $r \in V, |V|=n$ deployed at random starting locations in an unknown environment. 
The connectivity of the robots is expressed by a graph $G=(V,E)$, and a robot $i$ can share data with their immediate neighbors $\mathcal{N}_i = \{j \in V | (i,j) \in E \}$.
The robots must find and navigate toward the frontier position on their locally built map as a standard map exploration strategy. To accomplish this efficiently, a robot must decide which frontier to navigate after leaving its current explored region and is expected to reduce the number of steps to take and the size of data exchanges between robots.

Robots only share updated Q-value and the newly explored frontier with other robots in its range (neighbor set). Each robot keeps track of its local and shared frontiers to avoid re-exploration. Robots continue to generate local maps and share the newly developed map only when asked by other robots in case of the already explored frontier. Robots who cannot find new frontiers merge their local map with a map received from peers to build a global map using the feature similarity-based map merging technique \cite{mangelson2018pairwise}. The robot's decisions regarding an action plan are based on the shared information and Q-learning computation. 
The whole procedure concerning robot $i$ can be visualized in Fig.~\ref{fig:system-architecture}.

\subsection{Q-learning}
Markov's decision processes frequently model the robot's interaction with the environment. A robot's state is $(x,y,\theta, active/inactive)$ in a global frame. 
Robots are localized and initialized in a global frame, and positions are known concerning virtually defined bounded regions that can be expanded based on exploration requirements.
We consider the frontier's position as states for exploration by applying efficient frontier detection \cite{keidar2014efficient}. \textcolor{\revision}{We consider the current frontier's position as the current state $s_t$ and the newly explored frontier as the next state $s_{t+1}$}. A robot can transition from state $s_t \in S$ to state $s_{t+1}$ due to acting $a_t \in A$ based on its state at time $t$. Robot action $a_t$ to reach $s_{t+1}$ from $s_t$ can be determined using discrete-time Hopfield function \cite{uykan2019working}. 
The transition probability is defined as $T: S \times A \times S \rightarrow [0, 1]$. The robot will receive a reward for each action using a reward function $R: S \times A \times S \rightarrow R$ specific to the task. The robot will have learned the course of action to take in each state and will be able to maximize the reward of the entire interaction process. 

In Q-learning, all possible states and actions are created using the Q table, which then updates each value through iterative learning. The robot then chooses the best course of action for each state based on the values in the table. This approach is frequently utilized in path planning, chess, card games, and other activities.






\textbf{Assumptions:} For simplicity, we assume a flat ground terrain environment for exploration.
\begin{itemize}
    \item The robots have omnidirectional sensors to detect obstacle boundaries within a maximum sensing range $r_s$.
    \item  {\color{\revision} Each robot has a communication range, $r_c >> r_s$ forming a neighbor set $\mathcal{N}_i$ with constant connectivity to send and receive information about its relative position.}
    {\item The connectivity graph $G$ is connected throughout exploration, which is practical to achieve in a multi-robot application. Since the proposed solution is distributed, it ensures maximum coverage even in partial disconnectivity, but at the cost of increased re-exploration.}
\end{itemize}

Here, we introduce CQLite as a distributed method for robot $i$, which is now at state $s_t$ at time $t$ and selects the following state as $s_{t+1}$ to explore independently. \textcolor{\revision}{CQLite begins with an empty Q-table and updates its table values as exploration proceeds; CQLite aims to achieve fast convergence based on the optimized Q-learning and reward function.}
Finding the action $a$ that maximizes the Q-value for a specific state $s$ is the goal of the maximum optimization function for Q-learning, i.e., 
    $a^*  = \argmax_a Q_i(s,a) $, 
where $a^*$ is the optimal action for a given state $s$.  

The Q-learning algorithm updates the Q value as 
\begin{align}
& {Q_{i,t + 1}}\left( {{s_t},{a_t}} \right) \nonumber \\
& = (1 - \alpha ){Q_{i,t}}\left( {{s_t},{a_t}} \right) + \alpha \left[ {{r_{i,t}} + \gamma {Q_{i,t}}\left( {{s_{t+1}},{a^*}} \right)} \right],
\label{eqn: q-value}
\end{align}
where $r_{i,t}$ is the reward received for taking action $a$ in state $s_t$.
The $\alpha \in (0, 1]$ controls the balance between the coverage and delay, and $\gamma$ is the discount factor to prioritize present vs. future rewards.
This optimization function is used in the action selection step of the Q-learning, where the agent selects the action that maximizes its expected future reward.

The objective of the CQLite is to perform maximum coverage in less time and avoid overlapping exploration, which can be numerically defined as 
\begin{align}
    max_\pi \{P_a^\pi (t) - \lambda_i E_t(a|\pi)\} ,
    \label{eqn:objective}
\end{align}
where $P_a^\pi(t)$ is the probability to cover the { unexplored} region using for action $a$ using policy $\pi$ at time $t$,  $E_t(a|\pi)$ is estimated time to reach the state $s_t$ by taking action $a$ at time $t$ in policy $\pi$ and $\lambda_i$ is the cost associated with each step taken by robot $i$. We have a vector $path$ extracted by containing position waypoints connecting $s_t$ to $s_{t+1}$ associated with $a$ \cite{li2020faster}. For each dimension of $path$ at each control instant $t=t_j$, we first compute the velocity command as:
\begin{align}
    v_{t_j} = K_p \cdot e_{t_j} + K_I \sum_{t=t_0}^{t_j}(e_{t_j}) ,
\end{align}
where $e_{t_j} = s_{t,j} - s_{t,j-1}$ represents the instantaneous error between the intermediate states associated with action $a$ (i.e., the feedback) at time $t= t_j$. Further, $K_p$ and $K_I$ are the so-called proportional and integral gains of the motor controller that regulate the contributions of the corrections induced by the actual error and the error accumulated over time, respectively. 
{These constant values determination is based on the motion constraints of our differential drive robots as discussed by Li et al. \cite{li2020faster}. They can be different based on the robot's physical and motion characteristics.} In our case, we predetermined values of $K_P$ and $K_I$ as 2 and 0.5, respectively. Now we apply simple kinematic to find the estimate $E_t(a|\pi)$ as:
\begin{equation}
        E_t(a|\pi) = \sum_{j=1}^{m} \frac{(e_{t_j})^2}{v_{t_j}} ,
\end{equation}
\textcolor{\revision}{where $m = |\mathcal{N}_i|$ is the number of intermediate neighbors of the robot $i$.}
To avoid the exploration of an already explored region for state $s_t$, we determine $P(s_t \cap ES_t)$ as:
\begin{equation}
    P(s_t \cap ES_t) = \sum_{j=1}^{m} \frac{P(s_t \cap es_j)}{m} ,
\end{equation}
where \textcolor{\revision}{$ES_t$ is the set of explored states at time $t$}, $es_j \in ES_t$, and $m$ is the number of explored states and overlap probability of each explored state in $ES_t$ can be determined as:
\begin{align}
P(s_t \cap es_j)=
\begin{cases}
1 & dist(s_t,es_j)\leq r_{i,s} \\
0 & dist(s_t,es_j) > r_{i,s} \\
\end{cases}
\end{align}
At each discrete time step $t$, the robot $i$ acquires an observation $s_t$ from the environment, selects a corresponding action $a_t$, then receives feedback from the environment in the form of a reward $r_{it}=R(s_t,a_t)$ as shown below:
\begin{align}
r_{it} =
\begin{cases}
-\lambda_i & s_t \in ES_t \\
\lambda_i - Q_{i,t} + \rho(1 - P(s_t \cap ES_t)) + 
\sigma r_{i,c} & s_t \notin ES_t \
\end{cases}
\label{eqn:reward}
\end{align}

Where $P(s_t \cap ES_t)$ is the probability of overlap between the current state $s_t$, and the already explored states $ES_t$ by $robot_i$ and other robots, $\rho$ is a scaling factor that controls the importance of minimizing the overlap, $r_c$ is the communication range, and $\sigma$ is the scaling factor that determines the importance of maximizing the communication range. {$\sigma$ depends upon the robot's sensing capabilities and makes the reward function modular for heterogeneous robots with different sensing capabilities \cite{ov2020impact}.}
Then the state information is updated $s_{t+1}$. The goal of the RL is to select policy $\pi$ that maximizes the discounted sum of future rewards, i.e., $Q_\pi(s_1) = \sum_\tau^ {t=1}\gamma^t R(s_t,a_t)$, which according to the Bellman optimality principle satisfies.

The reward function in Eq.~\eqref{eqn:reward} produces a negative reward whenever the agent has looped back, and the calculated reward is based on the step-cost, Q-value, probability of overlap, and scaling factor otherwise.

\textbf{Multi-Robot Lite Cooperation:} We reduce the communication overhead amongst individual exploration-capable robots through a distributed approach, allowing each robot to make independent decisions based on local information and with little interaction from other robots. In our lite version of Q-learning, only the current state and Q-value are communicated amongst nearby robots to encourage cooperation. When another robot receives the information, it will update the received Q value in its Q table and update the local map. We develop a discovery approach based on the distance between simulated robots to replicate the network range in which we only share the current position of a robot $i$, its Q-Value for each direction, and mark the current situation as explored to avoid repetitive exploration.

\subsection{Exploration Strategy} 
Robots create a global Q table for each cell and action after searching the map and experiencing several experiences. The Q table is then turned into a weighted graph $G=(\mathcal{S}, \mathcal{E}, \mathcal{C})$, where $\mathcal{S} =\{s_1,s_2,...,s_n\}$ denotes the set of states, and $\mathcal{E}\in |\mathcal{S}|\times|\mathcal{S}|$ signifies the set of edges whose elements indicate whether or not a path exists between the center points of each pair of states. {It is assumed that robots do not exchange nodes during exploration, and Voronoi boundaries are fixed.} Furthermore, $\mathcal{C}$ is the weight matrix indicating the edge metric cost. 
The primary goal of discovering this study's reduced graph and significant states is to optimally disperse robots over the coverage region by minimizing the relevant cost function. Because robots move at varying speeds, we formulate the cost as a function of the defined traveling time as 
    $t_{(s_{p_i},s_q)}= \frac{d_{(s_{p_i},s_q)}}{v_i},$
where $v_i$ is the $i^{th}$ robot's speed, and $d_{(s _{p_i},s_q)}\in \mathcal{C}$ is the Euclidean distance between the $i^{th}$ robot's current state $p_i$ and state $q$. Furthermore, knowing the optimal path from state $p_i$ to state $s$, each robot's overall optimal traveling time is the sum of the trip times (costs) from state $p_i$ to state $s$. This study's shortest path between each pair of states is computed using the A* algorithm. Then the total time $\tau$ is calculated by knowing path $\mathcal{P} = \{p_1,p_2,...,p_n\}$ as
\begin{align}
\tau_{(S_p,S_q)} = {t_{(s_{p},s_{p_1)}} +t_{(s_{p_1},s_{p_2})}+ ... + t_{(s_{p_n},s_q)}} .
\label{eqn:tau}
\end{align}

After determining the shortest time between each pair of states, the field is partitioned into M Voronoi subgraphs $g_{r_i}$ for $i \in \{1, 2, ..., M\} $ to distribute work proportionally among $M$ robots. To that aim, the ideal Voronoi diagram $g_{r_i}$ for $i^{th}$ robot, according to Lloyd's algorithm, is a split of the area determined as:
\begin{align}    
    g_{r_i} = \{s_q \in S | \tau_{(s_{p_i},s_q)}\leq {\tau_{(s_{p_j},s_q)}}, \forall i\neq j\} .
    \label{eqn:gsplit}
\end{align}
{Where $j$ is the other connected robot.}
The $i^{th}$ robot is responsible for covering the state $s$ (associated robot) in its sub-graph $g_{r_i}$ using the Voronoi partitioning result. The entire cost is then calculated as 
\begin{align}
    \color{\revision}
    \lambda_{i,(p,g_r)}= 
    {\sum_{q\in g_{r_i}} \tau_{(s_{p_j},s_q)} \phi_q} ,
    \color{black}
    \label{eqn:lambda}
\end{align}
where $\phi_q$ is the priority value associated with state $s_q$. As the map turns into a graph, higher priority values are assigned to target states, while lower priority values are assigned to states far and already explored from the current state. The entire travel time (cost) will therefore be minimized, and an optimal solution will be obtained only when the current distance between the robot $i$ and the target state $s_q$, $d_{(s_{p_i},s_q)}$ converges to zero. Algorithm~\ref{alg:q-learning} provides the pseudocode description of CQLite for efficient map exploration implemented in a distributed manner on each robot $i$.

\begin{algorithm}[tb]
 \SetAlgoLined
 \color{\revision}
 Data: Reward matrix $R$, learning rate $\alpha$, discount factor $\gamma$, step cost $\lambda_i$\;
 Number of iterations: $t = 0$\;
 Initialize empty Q-table as $Q_{i}$\ for robot $i$;
 Initialize empty explored frontier list $ES_t$\;
 Generate a local map using range sensor\;
Initialize Explored Frontier Detected $s_d$ as 0\;
 \While{ $(t \leq t_{max})$ and $s_d$<2}
 {
    $S_t \rightarrow S_{start}$, step = 1\;
    Find new frontiers at new $ES_t$ and update $S_t$ using locally explored map\;
        $Q_{i,t}$ = list(0)\;
        \For{each frontier $s$ in $S_t$}{
            \eIf{$s$ not in $ES_t$}{
                Calculate Q-value as $q_{s}$ for actions $a_{s}$ to reach frontier $s$ using Eq.~\eqref{eqn: q-value}\;
                Append $Q_{i,t}(s, a_s)$\;
             }
             {
                Request for explored Maps\;
                Merge maps into local maps\;
                $s_d$ = $s_d$ + 1\;
             }
            }
        \If{$f_d<2$}{
            Set updated Q-value $q_{update}$ as $max_a(Q_{i,t}(s,a))$\;
            Update $Q_{i}$ with $q_{update}$\;
            Take action $a_t$ associated with $Q_{i}$\;
            Share $q_{update}$  with connected robots\;
            Receive Q-value for $ES_t$ from connected robots $j\in \mathcal{N}_i$\;
            Receive explored frontiers $es_{1:n-1}$ and update $ES_t$ with $s_t$ and $es_{1:n-1}$\;
            Set new state associated with $a_t$ as $S_t$\;
            Reset $s_d$ as 0\;
        }
    }
    \color{black}

 \caption{Distributed CQLite Exploration}
 \label{alg:q-learning}
 \end{algorithm}

\section{Theoretical Analysis}
We theoretically analyze the convergence, efficiency, and time complexity of our proposed CQLite for learning-based multi-robot map exploration.

\subsection{Convergence}
We analyze the convergence of the target Q value update function Eq.~\eqref{eqn: q-value}. We denote the error ratio $\delta_t = \frac{MSE(Q_t)}{E_t(a|\pi)}$, where $MSE(Q_t)$ is the calculated mean square error for Q-table at time $t$ and $E_t(a|\pi)$ is the average number of steps to cover the region by taking action $a$ at time $t$ for policy $\pi$.

\textbf{Theorem 1 (Convergence of Q-values):} Using Eq.~\eqref{eqn: q-value} for Q-value updates, then if $0\leq \delta_t\leq 1$, with probability $1-e$, we have the estimated time to reach a given state as:
\begin{align}
    E_t(a|\pi) \leq \omega E_1(a|\pi) + \sqrt{\frac{\ln(1/e)\sum_{i=0}^{t-1}\psi_i^2(\delta_{t-i:t})}{2}} 
    \label{eqn: theorem1}
\end{align}
Here, $\psi_i(\delta_{t-1:t})= \frac{\prod_{j=t-i}^{t-1}(j+\gamma\delta_j)}{\prod_{j=t-i}^{t}j}$, $\alpha_t= \frac{\prod_{j=1}^{t-1}(j+\gamma\delta_j)}{\prod_{j=2}^tj}$ and $\gamma = 0.95$.

\begin{proof}
Our analysis is derived based on the subsequent (synchronous) Q-learning. In contrast to the conventional  Q-learning, we swap out the current $Qt$ for the independent Q-function $Q'(s, a)$ for the target $Q_t(s_t, a_t)$ and note that if $Q'_t(s,a) = Q^*_{source}$, we know that $0\leq \delta_t\leq 1$.\\
First, we break down the update role into:
\begin{align*}
&{Q_{t}}\left( {{s_t},{a_t}} \right) \\
    & = \left(\frac{t-1}{t}\right) Q_{t-1}(s,a) + \frac{1}{t} \big( r_t + \gamma max_{a'} Q'_{t-1}(s',a')\\
  & + \gamma max_{a'} Q^*_{t-1}(s',a') - \gamma max_{a'} Q^*_{t-1}(s',a')\big )
\end{align*}
Let $\varepsilon_t(s,a)=Q_t(s,a)-Q^*(s,a)$ and \\  $\xi(s') =\gamma \times max_{a'} \left(Q^*_{t-1}(s',a')\right)$ then recall the definition of $\delta_t$, we will have $\varepsilon_t(s,a)$
\begin{align*}
    & \leq \frac{t-1}{t} \varepsilon_{t-1}(s,a) + \frac{1}{t} \big(  \xi(s') - E_{s'}\xi(s')\big) + \frac{1}{t} \gamma \delta_t E_{t-1}
\end{align*}
As we know $\varepsilon_t(s,a) \leq E_t$, by applying maximization and recursion of $E$, we will have:
\begin{align*}
   & E_t \leq \frac{t-1+ \gamma \delta_t}{t}E_{t-1} + \frac{1}{t} \big(\xi(s') - E_{s'}\xi(s')\big) \\
   & \leq  \frac{\prod_{j=1}^{t-1}(j+\gamma\delta_j)}{\prod_{j=2}^tj} E_1 + \sum_{i=1}^{t-1} \frac{\prod_{j=t-i}^{t-1}(j+\gamma\delta_j)}{\prod_{j=t-i}^{t}j}\\
   & \times \big(\xi(s') - E_{s'}\xi(s')\big)\\
   & = \alpha_t E_1 + \sum_{i=1}^{t-1} \psi_i(\delta) \big(\xi(s') - E_{s'}\xi(s')\big)
\end{align*}
According to weighted Hoeffding inequality \cite{duda2014novel}, with prob-
ability $1-e$, we can prove Eq~\eqref{eqn: theorem1} for Theorem 1.
\end{proof}

This convergence result demonstrates the influence of the error ratio on the convergence rate. In other words, learning will go more quickly for our chosen Q value update function. Even though CQLite shares only updated Q-value, it still achieves the required convergence and provides an optimal strategy for robots to explore the map efficiently. 


\subsection{CQLite Efficiency}
\textbf{Proposition 1 (Q-table Update Efficiency):} The CQLite Exploration method reduces the communication and computation cost for exploration by sharing and appending only updated Q-values and newly discovered frontiers to the local Q-table, which reduces communication and computation cost by $\frac{1}{n}$ than the cost of the SOTA approaches. Where $n$ is the total number of possible states (size of Q-table). 

\begin{proof}
The CQLite approach reduces the size of the Q-table and the amount of data that needs to be transmitted between robots by sharing and appending only the updated Q-values and recently found frontiers to the Q-table.
We prove this by comparing the data needs with that of the SOTA, where the full Q-table is shared between robots.
The shared Q-value for a given state-action combination $(i,j)$ in the Q-table will be $Q_{i,j}$. 

CQLite only updates Q-value once during the whole exploration, in contrast to SOTA as it updates each value in every iteration.
Compared to sharing and updating the whole Q-table, the communication and computing costs are decreased by $\frac{1}{n}$.
The update cost of CQLite for Q-table with size $n$ is $n$, but the update cost of SOTA is $n^2$ ; hence the cost reduction relation case be written as:
\begin{align*}
    C_{i,CQLite} = \frac{1}{n} C_{i,SOTA},
\end{align*} 
where $C_{i,SOTA}$ is the communication and calculation cost of updating and sharing the whole Q-table in SOTA exploration techniques, and $C_{i,CQLite}$ is the communication and computation cost of the CQLite Exploration method.

 To further determine the effectiveness of the Q-table update in CQLite, the cost of sending the matrix $Q$ over a network can be used to indicate the cost of sharing and updating the whole Q-table and can be stated as follows:
\begin{align*}
C_{i,SOTA} = \kappa \cdot \sum\limits_{j=1}^{n}|Q_{i,j}| ,
\end{align*}
where $|Q_{i,n}|$ is the absolute value of the Q-value of state $j$  for robot $i$, and $\alpha$ is a constant that denotes the cost of sending one unit of data across the network. {SOTA requires all Q-values for policy determination; hence all Q-values are shared to update Q-table in every iteration.}

The CQLite Exploration approach reduces the size of the matrix and the quantity of data that needs to be transferred by sharing and appending only the updated Q-values and newly found frontiers. Let $Q'$ be the updated matrix that only includes the new frontiers and updated Q-values. This modified matrix's transmission cost can be expressed as
\begin{align*}
C_{i,CQLite} = \kappa \cdot \sum\limits_{j=1}^{n}|Q'_{i,j}| .
\end{align*}

Since $Q'$ is a subset of $Q$, it can be concluded that $\sum\limits_{j=1}^{l} |Q'_{i,j}| \le \sum\limits_{j=1}^{l} |Q_{i,j}|$, and therefore:
\begin{align*}
C_{i,CQLite} \le \frac{1}{n} C_{i,SOTA}
\end{align*}

This proves that the CQLite exploration approach is more efficient regarding Q-table updating than the SOTA exploration methods like RRT and DRL.
\end{proof}

\textbf{Proposition 2 (Mapping Efficiency):} CQLite performs map sharing and merging with the probability $P(s_t \cap ES_t)$, which requires $<<{iterations}$ compared to relevant SOTA exploration approaches (e.g., RRT and DRL) for maximum exploration.
Here, $P(s_t \cap ES_t)$ is the probability of overlap between the current state $s_t$, and the already explored states $ES_t$ by $robot_i$ and other robots $iterations$ is the total number of iterations carried out by the algorithm.

\begin{proof}
The probability of overlap $P(s_t \cap ES_t)$ between the current state $s_t$ of robot $i$ and the previously explored states $ES_t$ by other robots is used to determine if map sharing and merging will take place in the CQLite Exploration technique. This map merging and sharing aims to reduce the number of iterations and steps the algorithm must perform.

As part of the CQLite Exploration approach, the algorithm updates the map by combining shared maps regularly as follows:
\begin{equation*}
f_{CQLite} = {P(s_t \cap ES_t)}\cdot {iterations}
\end{equation*}

Where $f_{CQLite}$ is the frequency of map merging carried out by the algorithm in the CQLite Exploration method, and ${iterations}$ is the mapping frequency of SOTA exploration methods like RRT and DRL.

The probability $P(s_t \cap ES_t)$ can be derived using Bayes' theorem as follows:
\begin{equation*}
P(s_t \cap ES_t) = P(ES_t \mid s_t) \cdot P(s_t)
\end{equation*}

Given the previously explored states $ES_t$, $P(ES_t \mid s_t)$ is the conditional probability of the current state $s_t$, and $P(s_t)$ is the probability of the current state $s_t$.

$P(s_t)$ can be represented as a uniform distribution over the state space, assuming that the exploration process is a random walk, with:
\begin{equation*}
P(s_t) = \frac{1}{n} ,
\end{equation*}
where the state space's overall state count is $n$.

The frequency of occurrence of the current state $s_t$ in the previously investigated states $ES_t$ can be used to estimate the conditional probability $P(ES_t | s_t)$. If the frequency with which the present state $s_t$ occurs in the previously studied states $ES_t$ is $f_{s_t}$, then:
\begin{equation*}
P(ES_t \mid s_t) = \frac{f_{s_t}}{n_e} ,
\end{equation*}
where $n_e$ is the total number of states in the already explored states $ES_t$.
Substituting the above expressions into the equation for $P(s_t \cap ES_t)$ gives:
\begin{equation*}
\begin{aligned}
P(s_t \cap ES_t) &= \frac{f_{s_t}}{n_e} \cdot \frac{1}{n} 
&= \frac{f_{s_t}}{nn_e} <<{iterations}
\end{aligned}
\end{equation*}

{For the total number of $iterations$ CQLite only updates the map for $\frac{f_{s_t}}{nn_e}$ times and $f_{s_t} < nn_e$ and $nn_e$ is equal to the $iterations$ in case of visiting each state at each $iteration$.} Hence, the above derivation proved that the CQLite method is more efficient as CQLite's update frequency (frequency of map merging) is  $<<{iterations}$  in map sharing and merging than SOTA exploration methods like RRT and DRL.
\end{proof}

Both propositions signify that the CQLite exploration method is more efficient in computation, communication, and mapping operations compared to the state-of-the-art RL-based multi-robot exploration approaches.

\subsection{Time Complexity} 
Assume that the grid factor is $k_g$ (resolution of the grid map on which the grid is divided) and that the target sub-map is $k\times l$ in size. The grid map's size is $k_g k\times k_g l$, and the total number of points is $k_g^2kl$. The operations to find Q-values must be carried out cyclically $k_g^2kl$ times.
$kl$ can calculate and represent the CQLite's state space size; however CQLite doesn't perform a merging and searching strategy at every iteration; hence the length of the Q-value table is significantly less than $(kl)$ through selecting the training process, and the computing complexity of the algorithm is considerably less than $O(kl)$.

\begin{figure}[t]
\centering
  \includegraphics[width=0.26\linewidth]{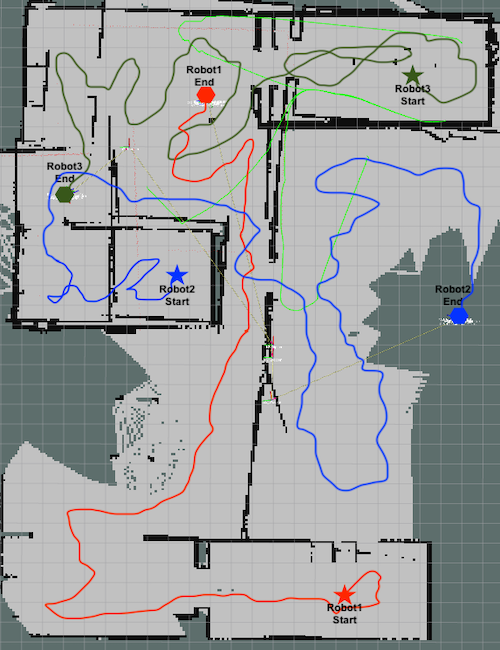}
  \label{fig:rrt_map_morerobots}
 \includegraphics[width=0.34\linewidth]{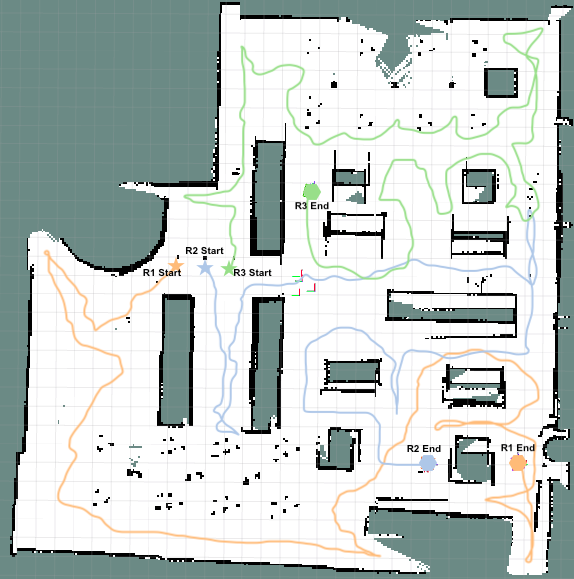}
 \label{fig:rrt_map}
  \includegraphics[width=0.33\linewidth]{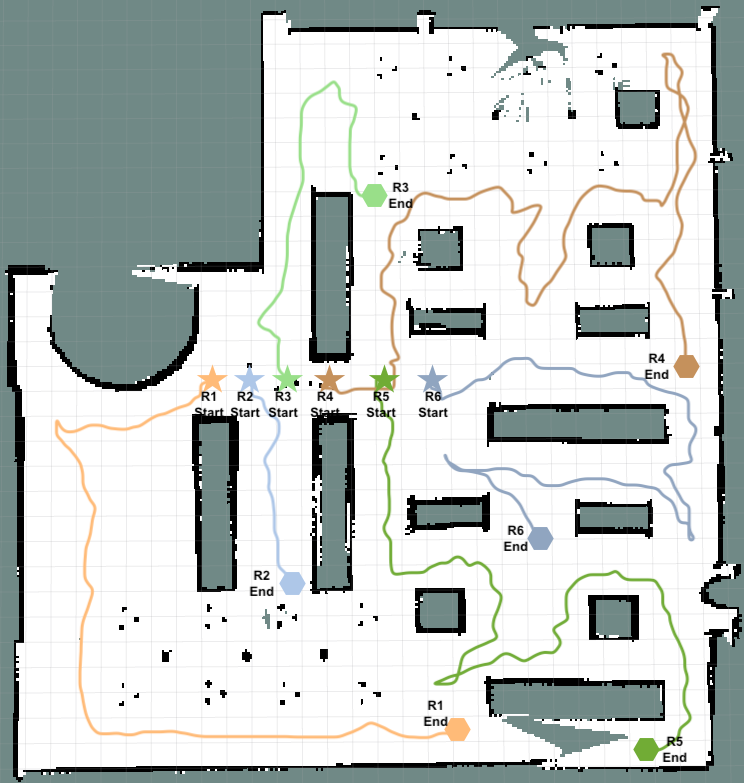}
 \label{fig:6robot_rrt_map}
  \vspace{2mm}
   \includegraphics[width=0.27\linewidth]{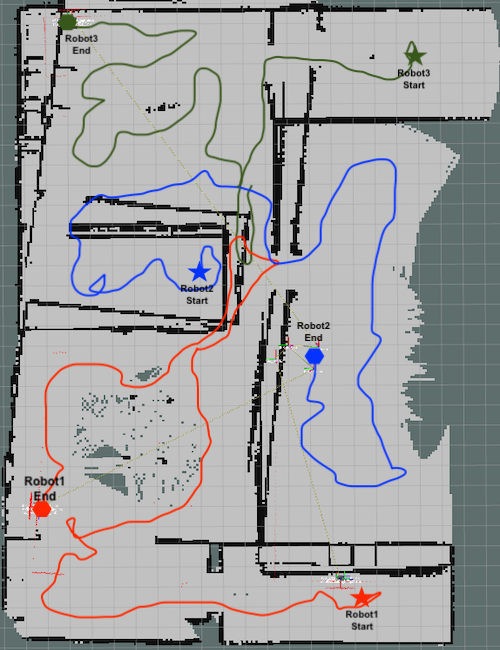}
 \label{fig:drl_map_morerobots}
 \includegraphics[width=0.33\linewidth]{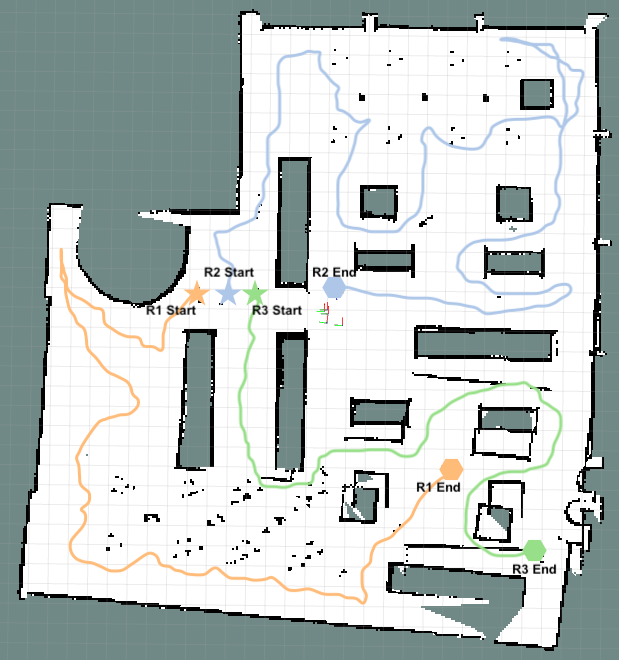}
 \label{fig:drl_map}
  \includegraphics[width=0.33\linewidth]{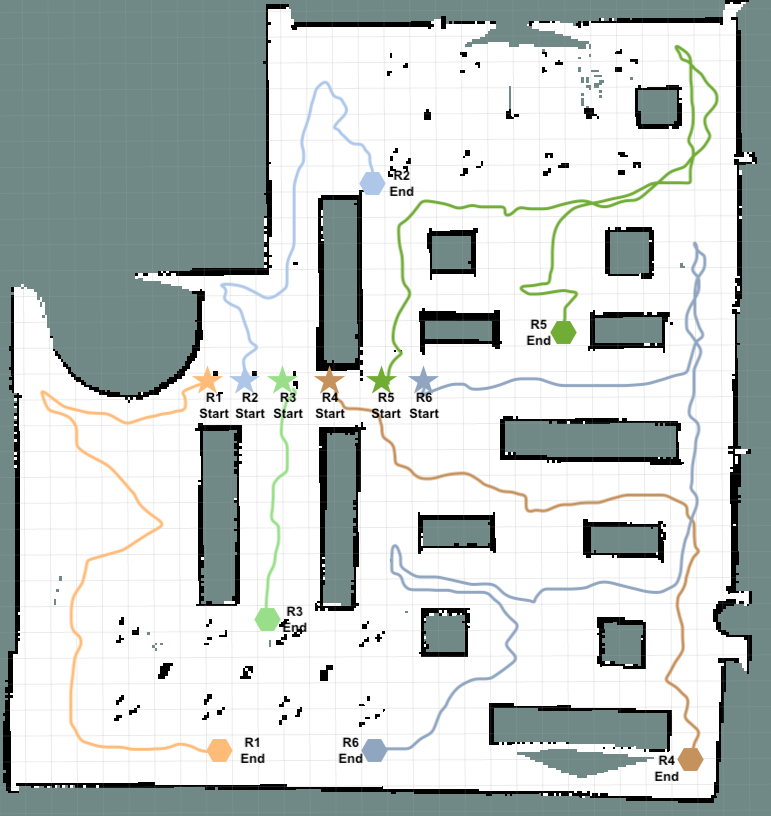}
 \label{fig:6robot_drl_map}

   \vspace{2mm}
    \includegraphics[width=0.27\linewidth]{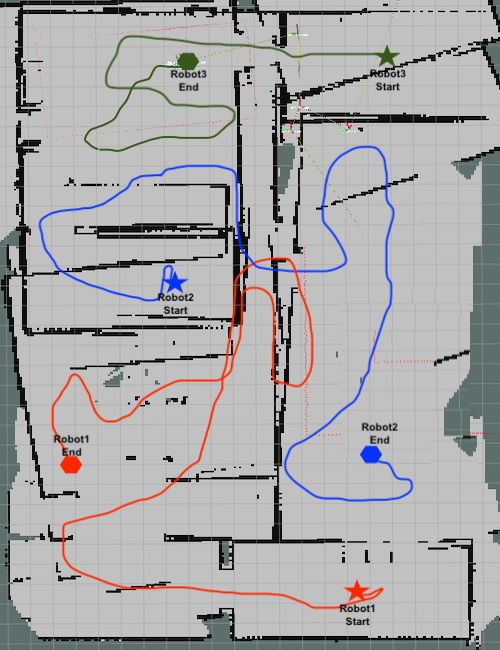}
 \label{fig:cqlite_map_house}
 \includegraphics[width=0.32\linewidth]{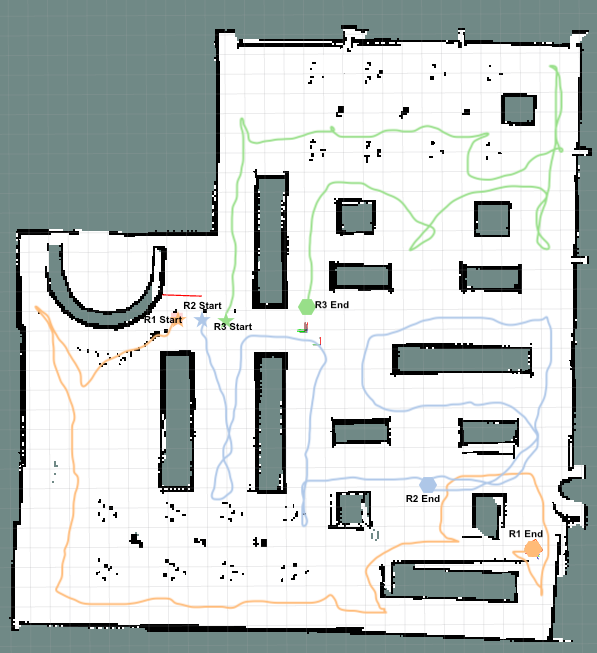}
 \label{fig:cqlite_map_bookstore3}
 \includegraphics[width=0.34\linewidth]{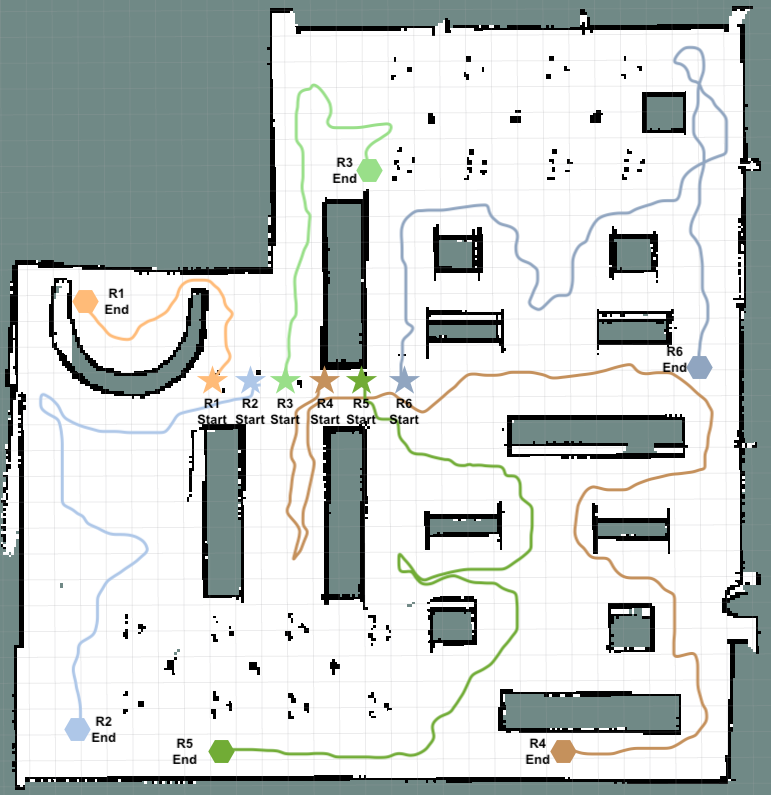}
 \label{fig:cqlite_map_bookstore6}
\caption{A depiction of the outcome in a sample trial. It shows the map generated by three robots in the house world (left column) and three and six robots in the bookstore world (canter and right column, respectively) created by the three compared approaches; RRT (top), DRL (center), and CQLite (bottom), with robots moving in a simulated House and Bookstore worlds along with the following trajectories, start and end locations.}
 \label{fig:explored-map}
\end{figure}

\section{Experimental Validation}
\label{experiments}
Turtlebot3 robots are used to carry out the exploration plan, implemented using the ROS framework. 
The open-source \textit{openslam-gmapping}\footnote{\url{https://openslam-org.github.io/gmapping.html}} technique of the ROS \textit{gmapping} package is used to create 2D maps. It uses odometer data and a particle filter method as its foundation. The local maps created by each robot are combined to create the global map. {Feature-based map merging\footnote{\url{http://wiki.ros.org/multirobot_map_merge}} is employed to merge maps when required.} Frame conversion between the local map frames is necessary for map merger. The coordinate transformation correlation between the robots must be calibrated before combining local maps. In the current work, the global frame is one robot's frame, and the relative positions and orientations of the robots are initialized to a known state.
The ROS \textit{movebase}\footnote{\url{https://github.com/ros-planning/navigation}} package allows the robot to move toward the goal point while securely avoiding barriers between robots. The Dijkstra algorithm for global path planning and the Dynamic Window Approach (DWA) for local dynamic obstacle avoidance are both implemented in this package. In this study, the units of time and distance are in seconds and meters, respectively.

 \begin{figure*}[ht]
 \includegraphics[width=0.33\linewidth]{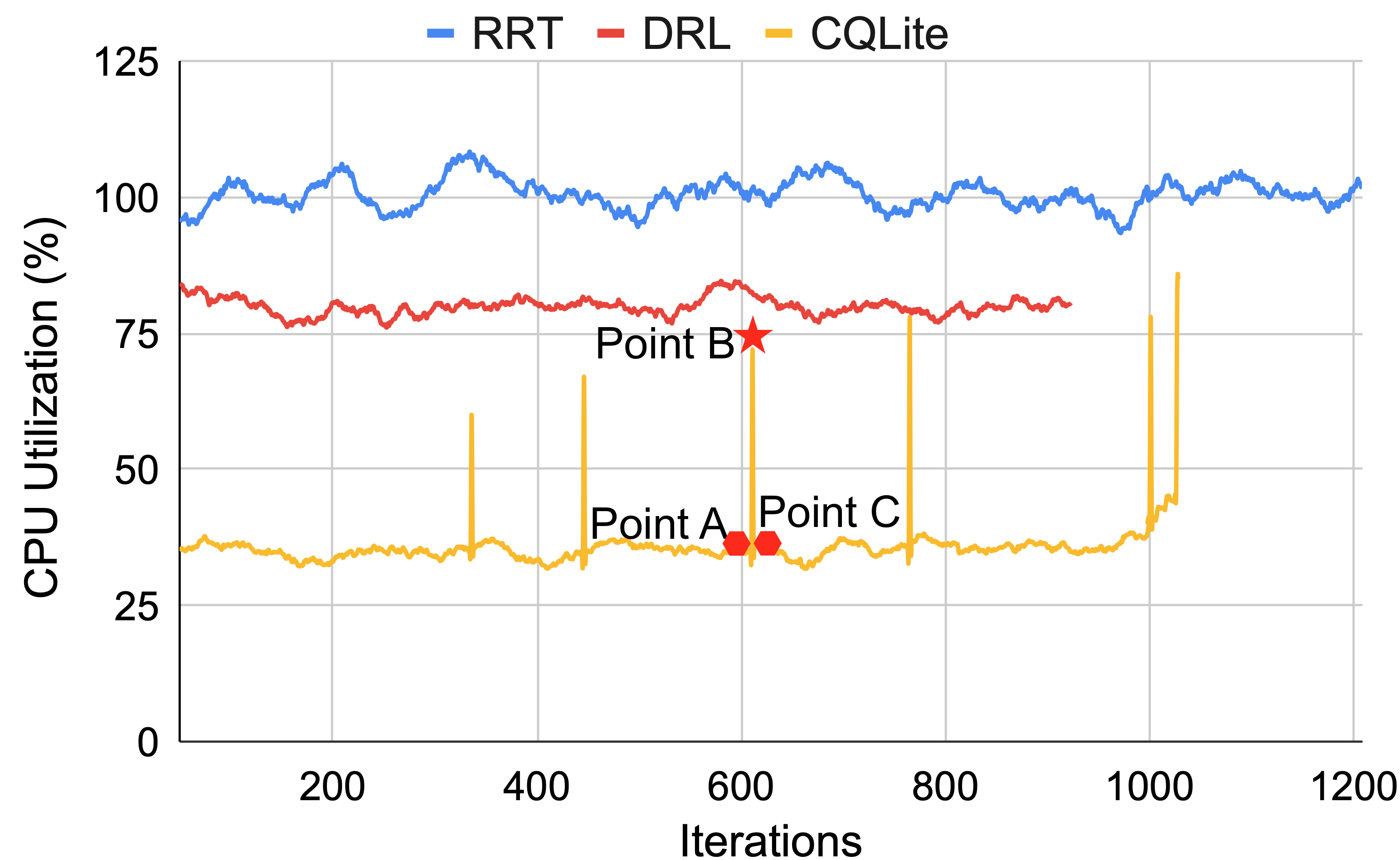}
 \label{fig:house_cpu_plot}
 \includegraphics[width=0.33\linewidth]{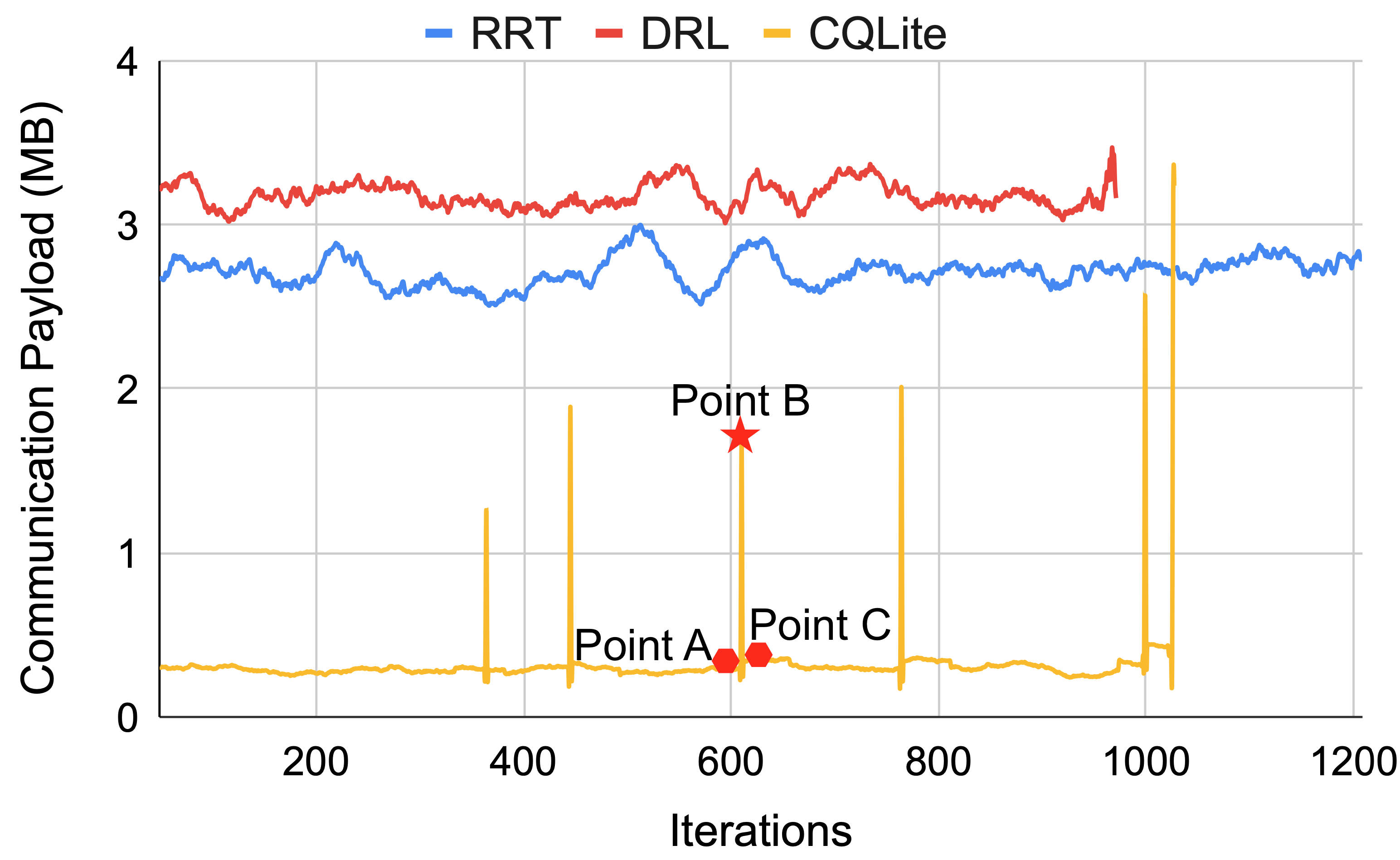}
  \label{fig:house_comm_plot}
\includegraphics[width=0.33\linewidth]{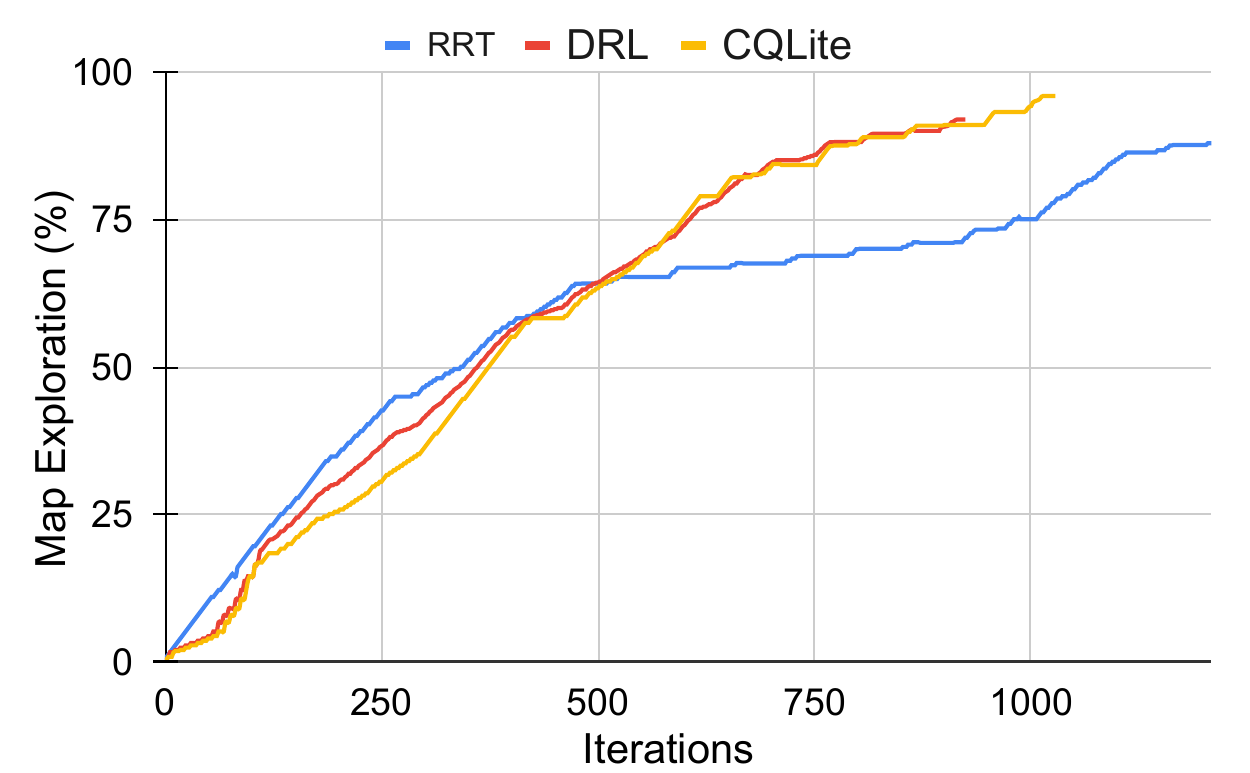}
  \label{fig:house_exploration}
  \includegraphics[width=0.33\linewidth]{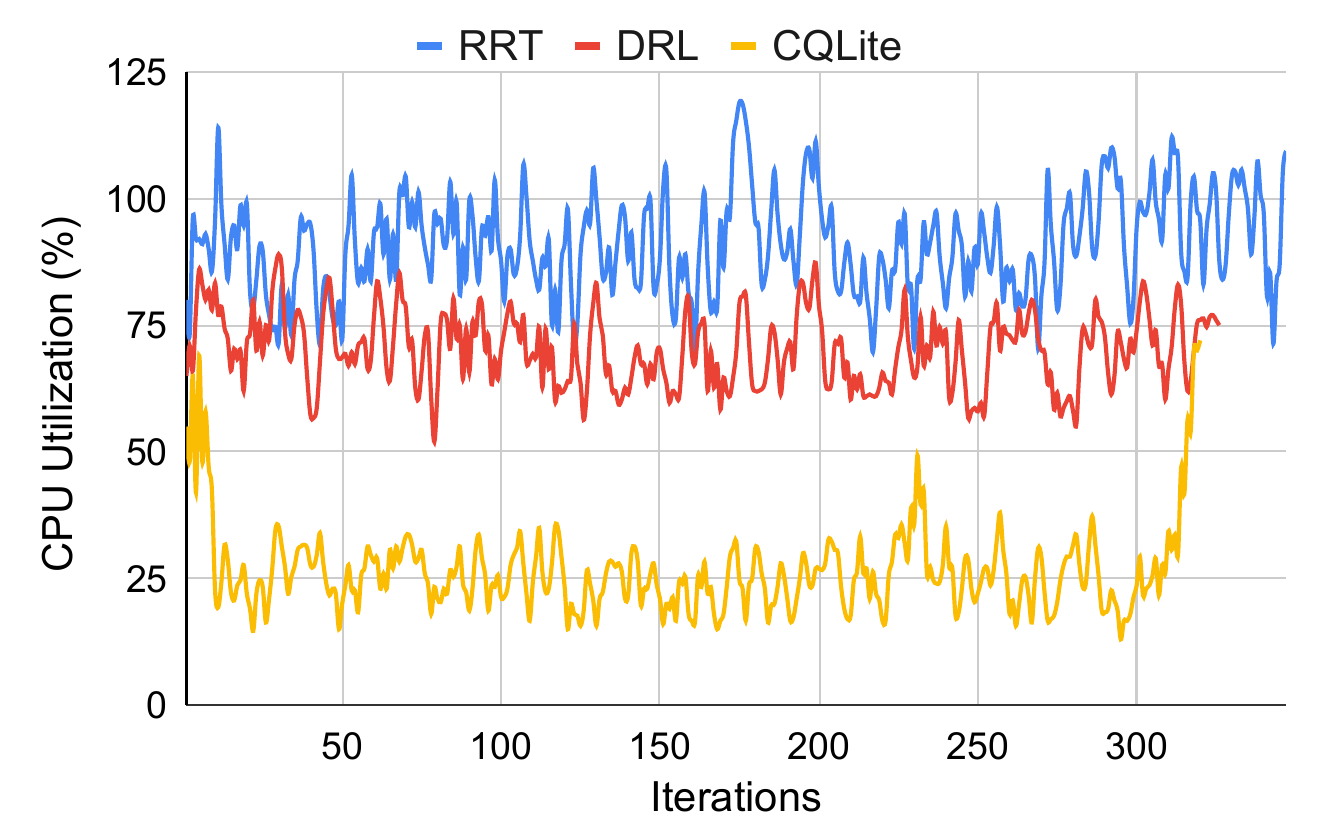}
 \includegraphics[width=0.33\linewidth]{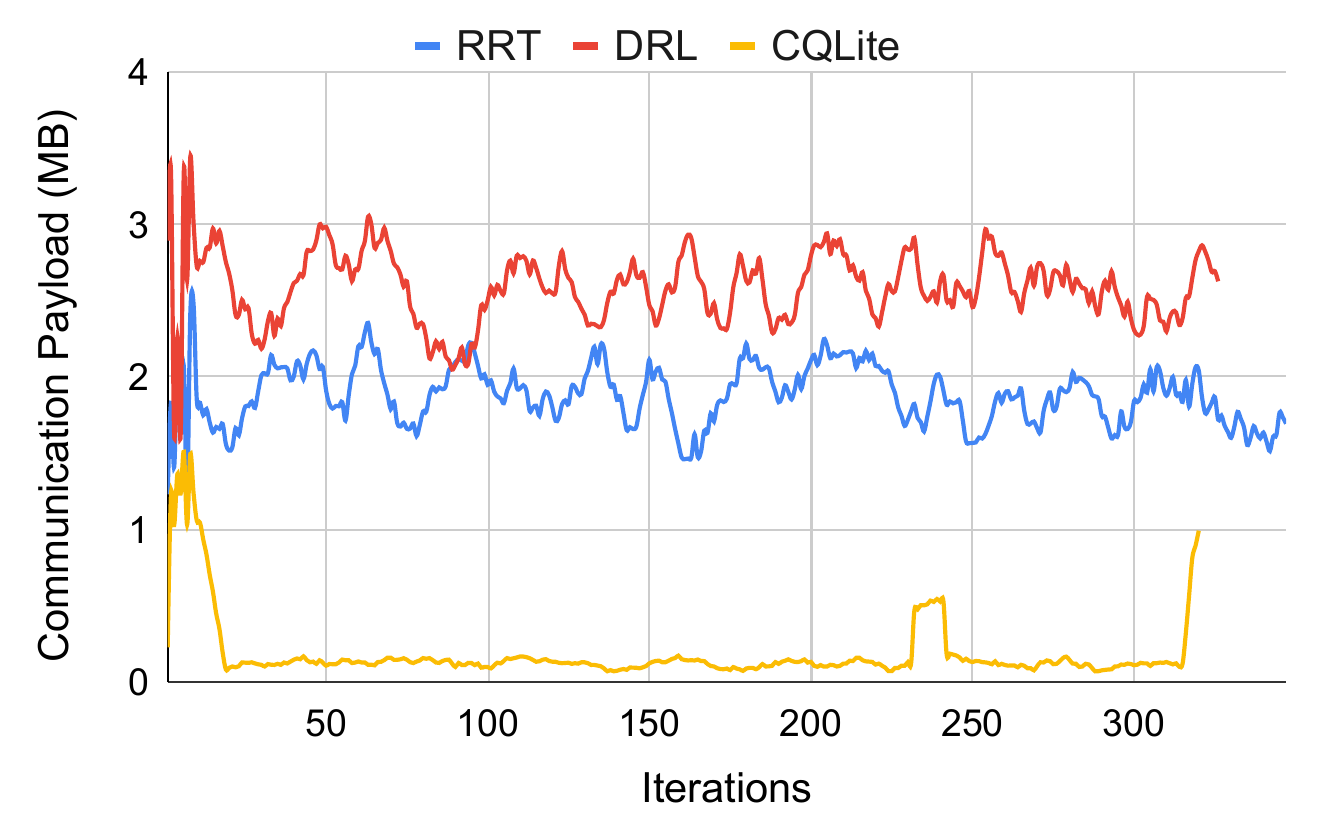}
\includegraphics[width=0.33\linewidth]{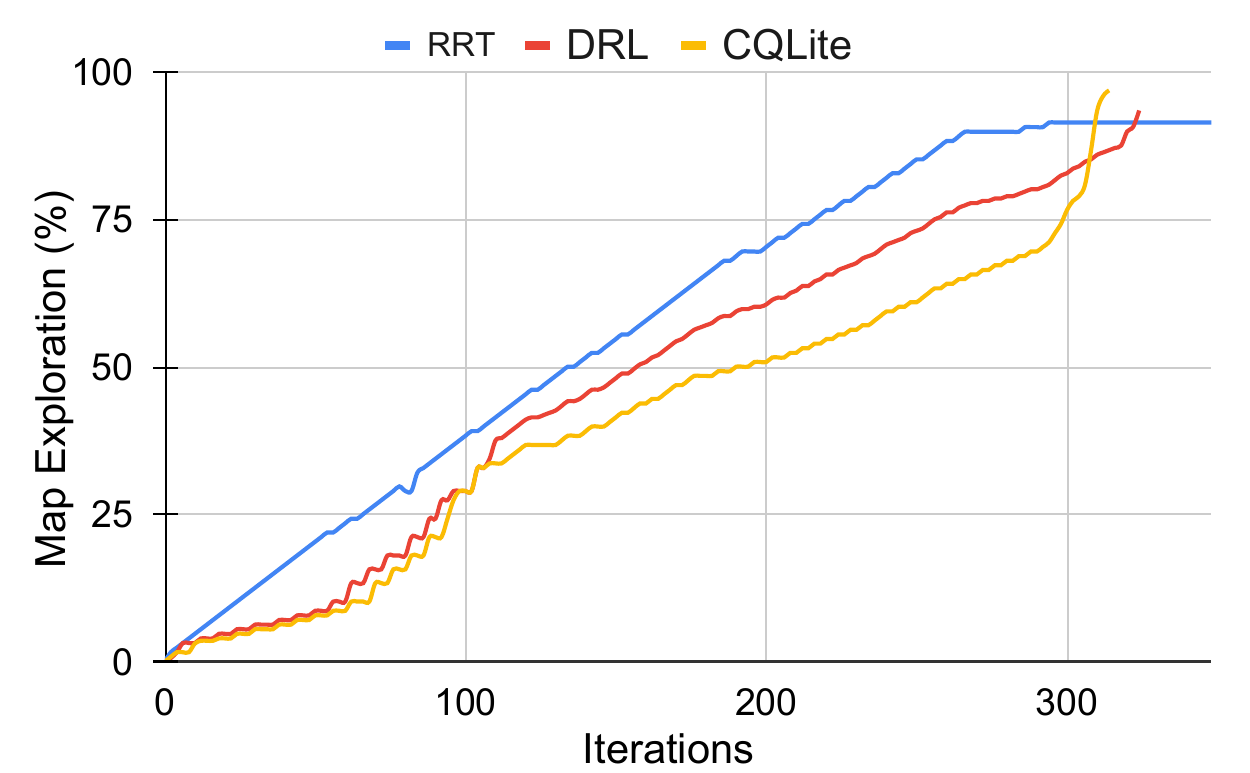}
 \includegraphics[width=0.33\linewidth]{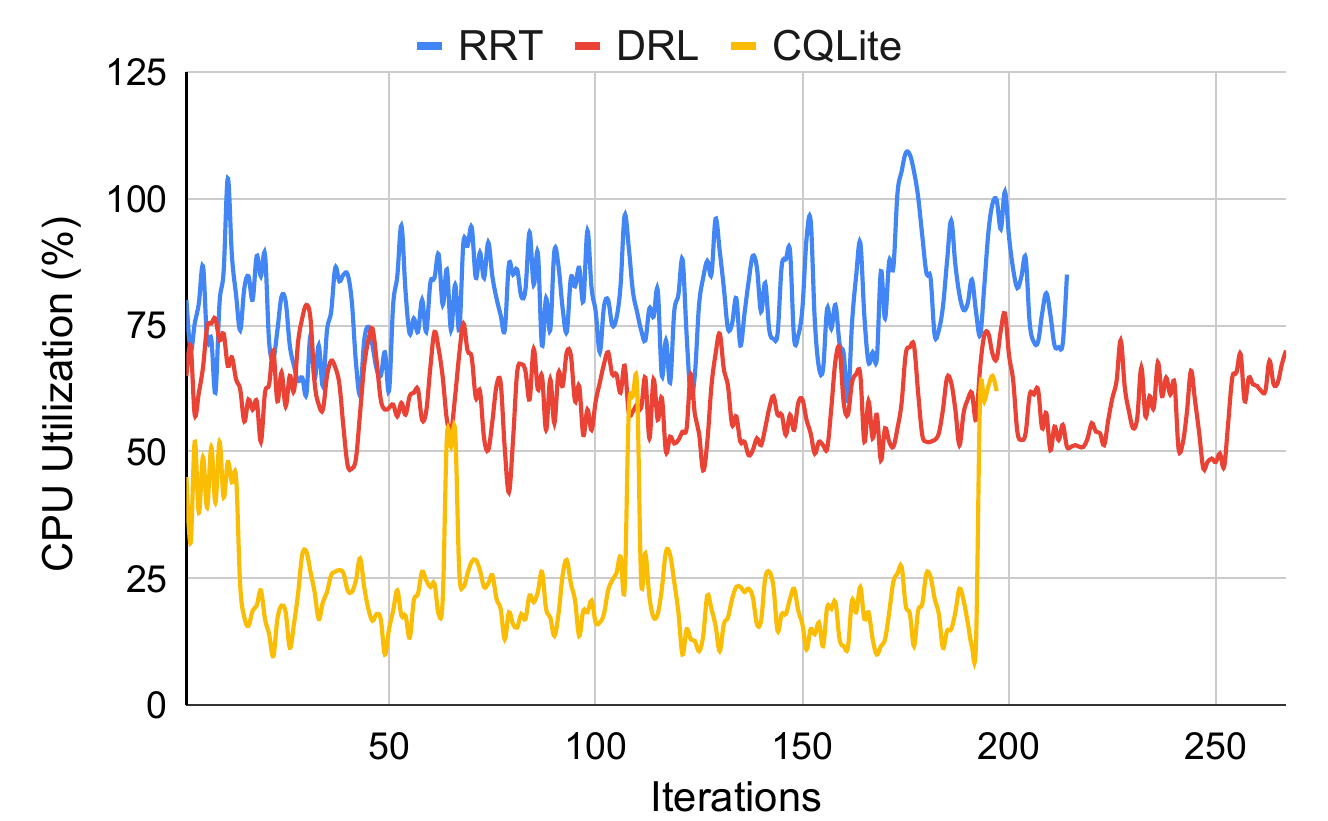}
 \includegraphics[width=0.33\linewidth]{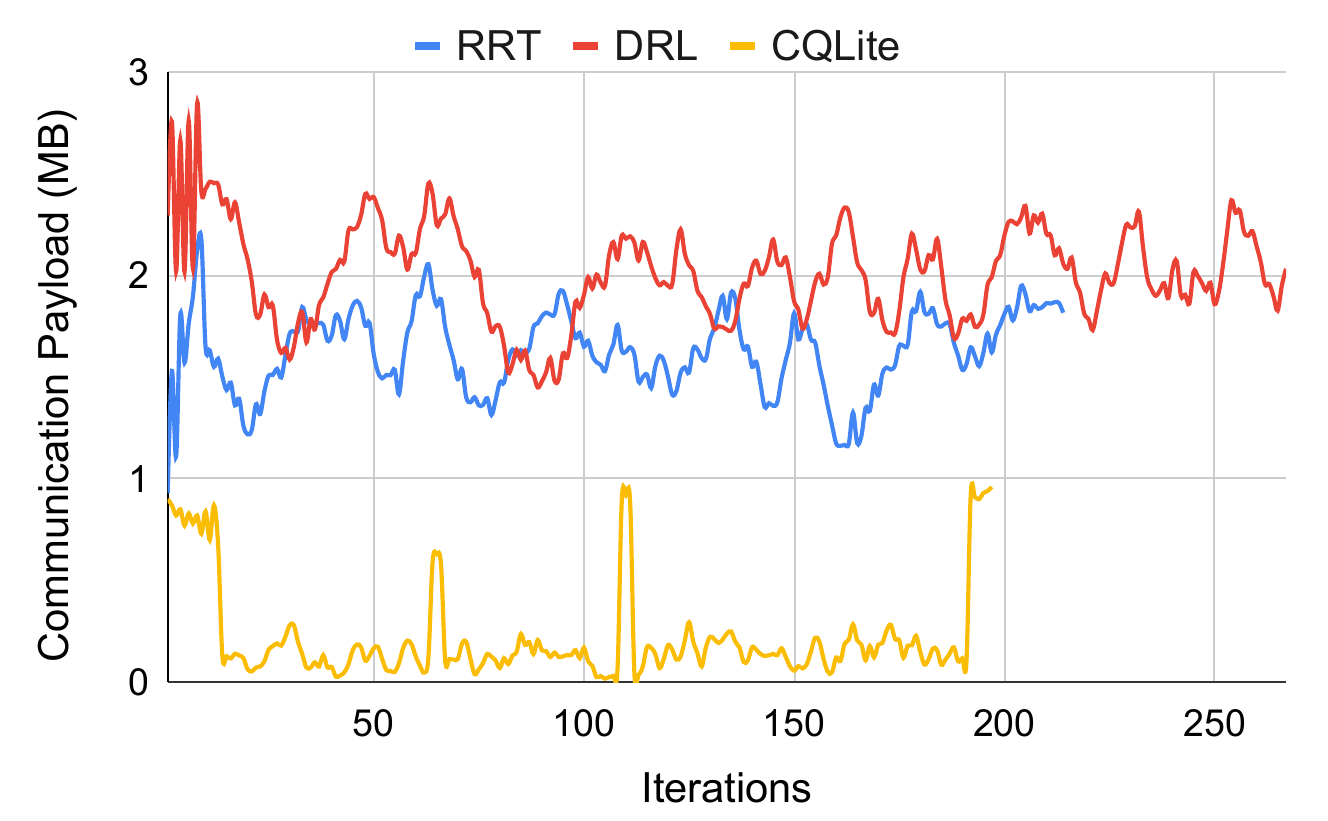}
\includegraphics[width=0.33\linewidth]{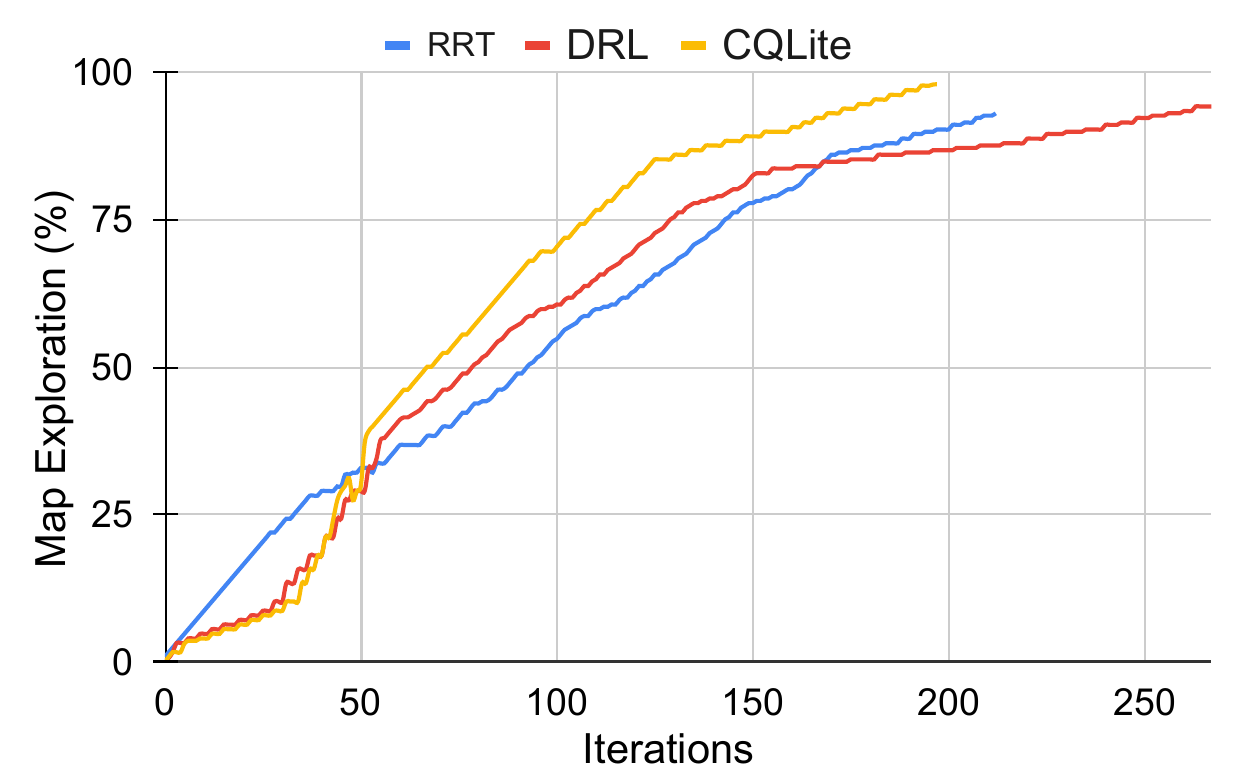}
\vspace{-6mm}
\caption{Computation \textbf{(Left)}, Communication \textbf{(Center)} cost, and Exploration over time \textbf{(Right)} comparison plot of CQLite with RRT and DRL approach in three Gazebo simulated world. Row-wise: \textbf{Top} 3 robots in house world, \textbf{Middle} 3 robots in bookstore world, and \textbf{Bottom} 6 robots in bookstore world.}
 \label{fig:comparison}
\end{figure*}

\begin{figure}[htb]
\centering
 \includegraphics[width=0.6\linewidth]{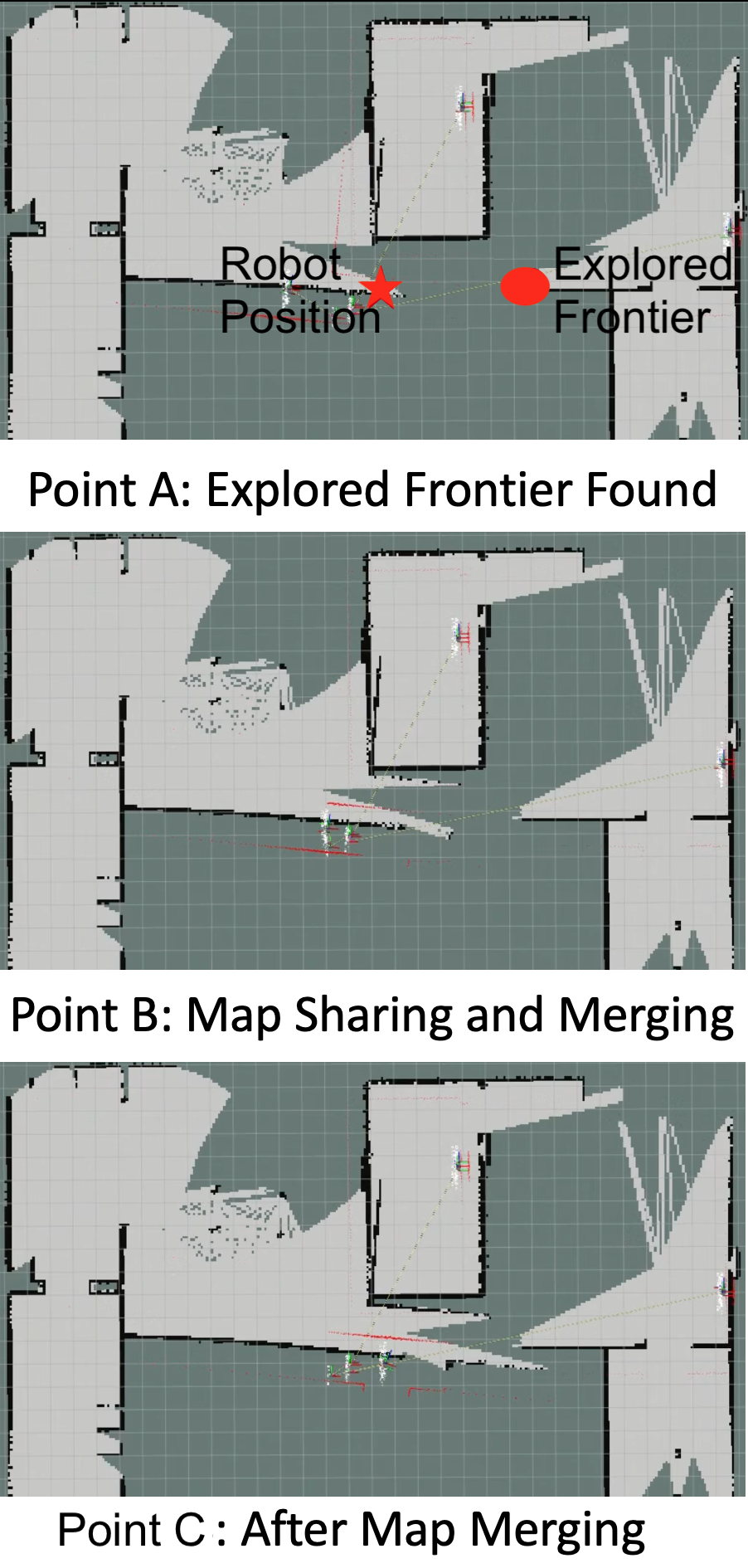}
\caption{Exploration map of the House world before, during, and after map sharing and merging corresponds to points (A, B, and C from Fig.~\ref{fig:comparison}) in Computation and Communication plots. Peaks demonstrate the request for map merging in CQLite for Computation and communication plots; RRT runs longer with persistent high communication and computational overhead but explores fewer regions than DRL and CQLite.}
 \label{fig:maps}
\end{figure}

\begin{table*}[ht]
\caption{Performance Results of RRT, DRL, and the proposed CQLite. $^*$ indicates the best performer.}
\label{table2}
\vspace{-2mm}
\resizebox{\linewidth}{!}{
\begin{tabular}{|c|c|c|c|c|c|c|c|c|c|}
\hline
\textbf{Evaluation parameters} & \multicolumn{3}{|c|}{\textbf{Three robots in house world}} & \multicolumn{3}{|c|}{\textbf{Three robots in bookstore world}} & \multicolumn{3}{|c|}{\textbf{Six robots in bookstore world}}\\
\cline{2-10}
& \textbf{RRT \cite{zhang2020rapidly}} & \textbf{DRL \cite{hu2020voronoi}} & \textbf{CQLite (ours)} &\textbf{RRT \cite{zhang2020rapidly}} & \textbf{DRL \cite{hu2020voronoi}} & \textbf{CQLite (ours)} &\textbf{RRT \cite{zhang2020rapidly}} & \textbf{DRL \cite{hu2020voronoi}} & \textbf{CQLite (ours)} \\
\hline
\textbf{Mapping Time (s)} & $1208\pm52$ & \textbf{924}$^*\pm67$ & $1029\pm59$& $347\pm32$ & $324\pm21$ & \textbf{317}$^*\pm19$& $212\pm18$ & $267\pm29$ & \textbf{197}$^*\pm13$\\
\hline
\textbf{Path Length (m)} & $592\pm11$ & $604\pm19$ & \textbf{543}$^*\pm9$ & $278\pm26$ & $235\pm29$ & \textbf{147}$^*\pm21$ & $223\pm12$ & $196\pm17$ & \textbf{121}$^*\pm11$ \\ 
\hline
\textbf{Exploration Percentage (\%)} & $87\pm4$ & $91\pm3$ & $\textbf{95}^*\pm3$  & $90\pm5$ & $93\pm2$ & $\textbf{97}^*\pm2$  & $93\pm4$ & $94\pm5$ & $\textbf{98}^*\pm2$ \\
\hline
\textbf{Overlap Percentage (\%)} & $51\pm5$ & $46\pm6$ & $\textbf{28}^*\pm2$ & $57\pm8$ & $51\pm9$ & $\textbf{31}^*\pm6$ & $47\pm7$ & $39\pm8$ & $\textbf{21}^*\pm6$ \\
\hline
 \textbf{MAP SSIM} &  $0.73\pm0.12$ & $0.89\pm0.08$ &  $\textbf{0.91}^*\pm0.06$  &  $0.68\pm0.21$ & $0.71\pm0.13$ &  $\textbf{0.89}^*\pm0.08$  &  $0.71\pm0.17$ & $0.73\pm0.15$ &  $\textbf{0.93}^*\pm0.10$ \\
\hline
\textbf{CPU Utilization (\%)} & $112\pm22$  & $79\pm18$  & $\textbf{42}^*\pm8$ & $97\pm18$  & $65\pm15$  & $\textbf{34}^*\pm9$ & $68\pm21$  & $47\pm16$  & $\textbf{26}^*\pm9$ \\
\hline
\textbf{RAM (MB)} & $824\pm19$  &  $1264\pm41$ & $\textbf{665}^*\pm24$  & $624\pm16$  &  $819\pm33$ & $\textbf{432}^*\pm21$  & $452\pm19$  &  $724\pm38$ & $\textbf{319}^*\pm18$  \\
\hline
\textbf{COM Payload (MB)} & $2.2\pm0.08$  &  $2.4\pm0.06$ & $\textbf{0.6}^*\pm0.02$   & $1.3\pm0.06$  &  $1.8\pm0.04$ & $\textbf{0.4}^*\pm0.01$   & $1.1\pm0.04$  &  $1.3\pm0.05$ & $\textbf{0.2}^*\pm0.01$  \\ 
\hline
\end{tabular}
}
\end{table*}

\subsection{Simulation Setup:} 
A closed simulation environment based on the ROS Gazebo simulator with two indoor template environments is used: the Gazebo's house world ($\approx$ 250$m^2$ area) and the Amazon AWS bookstore world ($\approx$ 100$m^2$ area). The robots may quickly finish the map exploration in a closed environment. Each robot has a laser scanner to gather data about its surroundings. 
The robots form a fully connected graph through a WiFi communication channel with a standard range of 40-60m. 
The robot's trajectory is determined based on the fusion of wheel odometry and laser scan information.

The following parameters are used in the experiments in the simulated environment. The laser scanner's range and $r_{i,s}$ are set to $15m$ and $1m$, respectively. Additionally, the robot's maximum linear and angular speeds in the simulation are set to $0.5 ms^{-1}$ and $\frac{\pi}{4} rads^{-1}$, respectively. The global detector's growth factor $\eta$ and the local detector's growth factor $\eta_1$ in the RRT detector are set to $5m$ and $3m$, respectively. The weight parameters, $\alpha = 0.6$, $\gamma = 0.95$ and $\lambda_i = 2$ for $1m$ distant step. 

{Each experiment was run for ten trials, with average observations reported.}
We evaluate the performance in the following three scenarios to validate the robustness and scalability of the proposed solution: 1) \textbf{3 robots} in the house world, 2)  \textbf{3 robots} in the bookstore world, and 3)  \textbf{6 robots} in the bookstore world.

\subsection{Evaluation Metrics}
The proposed CQLite and the methods put forward by RRT \cite{zhang2020rapidly} and DRL \cite{hu2020voronoi} are compared in our experiments. 
We use the below metrics for a comprehensive evaluation:
\begin{enumerate}
    \item \textbf{Mapping Time}: The amount of time spent mapping is a gauge of the efficiency of the exploration process; 
    \item \textbf{Path Length:} This term refers to the path length of all robot's trajectories combined until exploration converges. The entire trajectory length gives an idea of the robot's energy usage while subtly describing its investigation's effectiveness; 
    \item \textbf{Exploration Percentage:} The percentage of the generated map with time elapsed; 
    \item \textbf{Overlap Percentage:} The percentage of the overlap of the explored map with time elapsed; 
    \item \textbf{Map SSIM:} Structural similarity index measure of generated maps compared with ground truth map to measure map correctness;
    \item \textbf{CPU Utilization:} The maximum \% consumption of the processor of a robot throughout the trajectory; 
    \item \textbf{Memory Consumption (RAM):} The maximum occupied memory by the robot throughout the trajectory;
    \item \textbf{COM payload:} The size of the data communicated by a robot averaged over iterations.
\end{enumerate}

\section{Results and Discussion} 
We have reported each approach's average performance after ten trials in each condition to reduce the measurement noise and analyze the statistical details. 
A sample of the mapping outcomes of the compared approaches with the trajectories followed by three robots in the simulated environment is shown in Fig.~\ref{fig:explored-map} {and generated maps also delineate the map correctness.} The outcome should be stated considering the average mapping time, distance traveled, and mapping efficiency. Mapping efficiency is determined by comparing with the original map, and reported percentages are normalized with gazebo world dimensions.

Table~\ref{table2} provides a comparative analysis of different methods on all the performance metrics and the statistical data from the results. It also lists the theoretical (algorithmic) computational complexity. 
Figs.~\ref{fig:comparison} shows the comparison of the approaches in the three key performance metrics: computation, communication, and exploration. Fig.~\ref{fig:maps} shows a zoomed-in view of the mapping process at three different closely timed instances.

The proposed CQLite reliably outperforms other strategies on the key performance metrics. CQLite covers a larger area in less time, improving mapping efficiency by 10\% while traveling 22 fewer meters than RRT in the experiment. In three-robot scenarios, CQLite was more effective than DRL and RRT, with 9\% and 8\% shorter mapping times, respectively. Its path length was also less than DRL's by about 38\%. The advantages became even more apparent when the trial involved six robots. While the mapping time was around 26\% faster than DRL and 7\% faster than RRT for the bookhouse world, the path length was about 38\% shorter than with DRL. \textcolor{\revision}{However, the mapping time of DRL is 10\% better than CQLite for house world because DRL is trained for the world and has an optimized path for coverage.}

CQLite had an exploration percentage that was 4\% greater than DRL in the three-robot scenario. This advantage persisted in the six-robot case, where CQLite's exploration percentage was almost 4\% higher than DRL's while maintaining the lowest overlap percentage. The stability and effectiveness of CQLite in multi-robot exploration tasks are highlighted by these results from various experiments.

\begin{figure}[t]
\centering
 \includegraphics[width=0.98\linewidth]{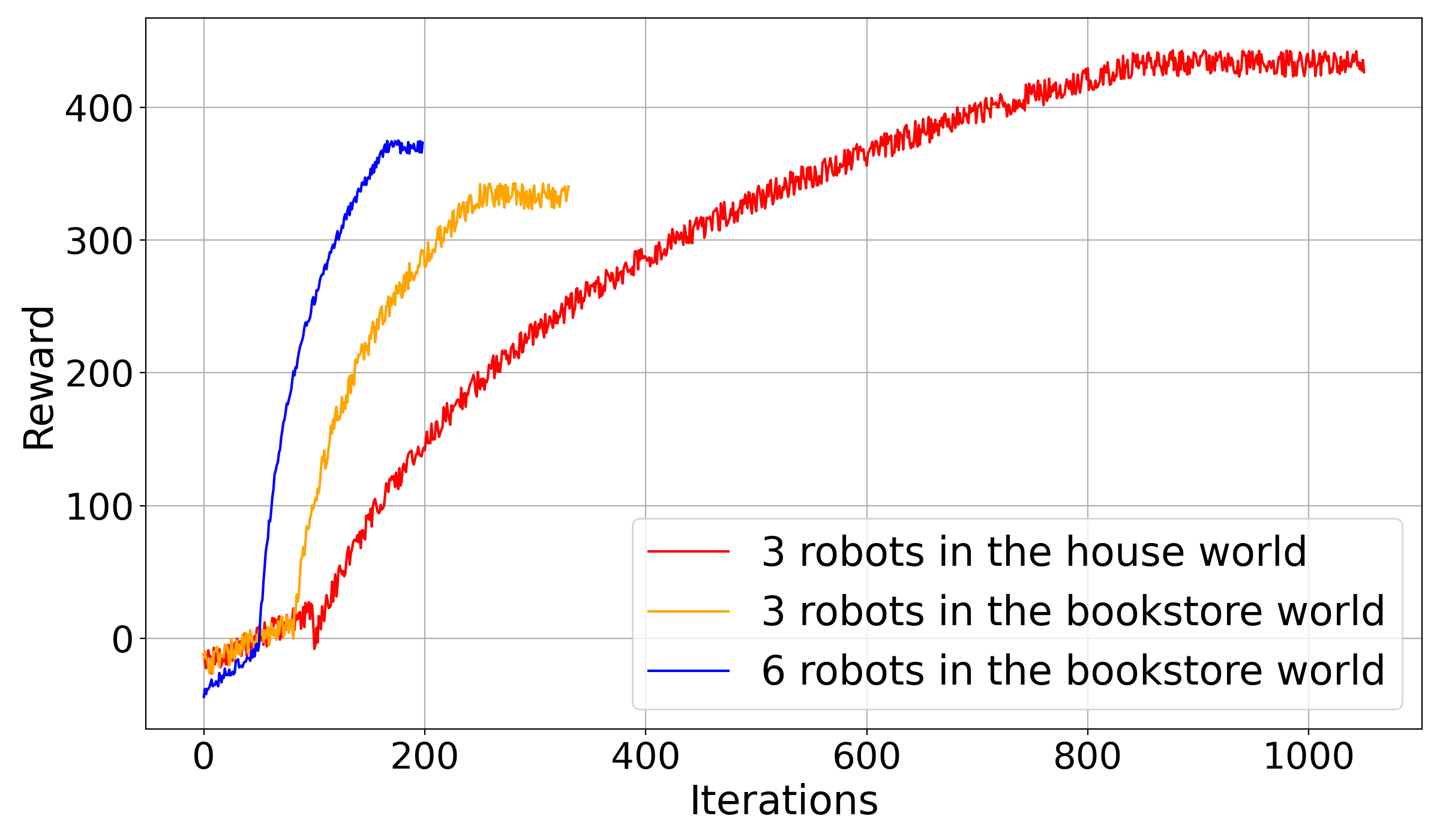}
 \caption{Reward progression with iterations for three experiments}
 \label{fig:reward}
\end{figure}

{\color{\revision}
In the reward convergence (Fig.~\ref{fig:reward}), the 3-robots in the house world show a steady progression, converging at around 850 iterations. The 3-robots in the bookstore demonstrate convergence near 280 iterations, while the 6-robots in the bookstore achieve stability quickest, by approximately 175 iterations. After these points, rewards in each experiment remain consistent, indicating optimal behavior within their respective environments. These convergence results imply the generalizability of the CQLite approach.
Further, it can be noted that the local convergence of Q-values (Theorem 1) within the robot's neighborhood set will lead to the group's global convergence when the graph is connected \cite{burgard2005coordinated,konda2020decentralized}.  
}

\color{black}
Communication-wise, CQLite's strategy is more effective. Contrary to RRT and DRL, which exchange locally explored maps continually. CQLite showed a significant reduction of more than 80\% in the communication payload (average data size) shared between the robots.
Notably, CQLite continues to explore at a constant rate even after reaching 60\% coverage, in contrast to RRT, which slows down. This dominance carries over into a real-life three-robot bookshop scenario, where it outperformed DRL and RRT regarding reduced mapping time and shorter journey distances.
Results have validated the practicality of CQLite by surpassing DRL and RRT in terms of most of the performance matrices in all scenarios. Further, demonstrating its efficacy and applicability on resource-constrained robots, CQLite maintained decreased RAM, CPU, and communication payload usage. CQLite demonstrates its power in managing a range of multi-robot exploration scenarios by offering improved map quality, as higher MAP SSIM ratings indicate. It is particularly appropriate in situations when there are significant communication and resource constraints. 

\begin{figure}[t]
    \centering
    \includegraphics[width=0.98\linewidth]{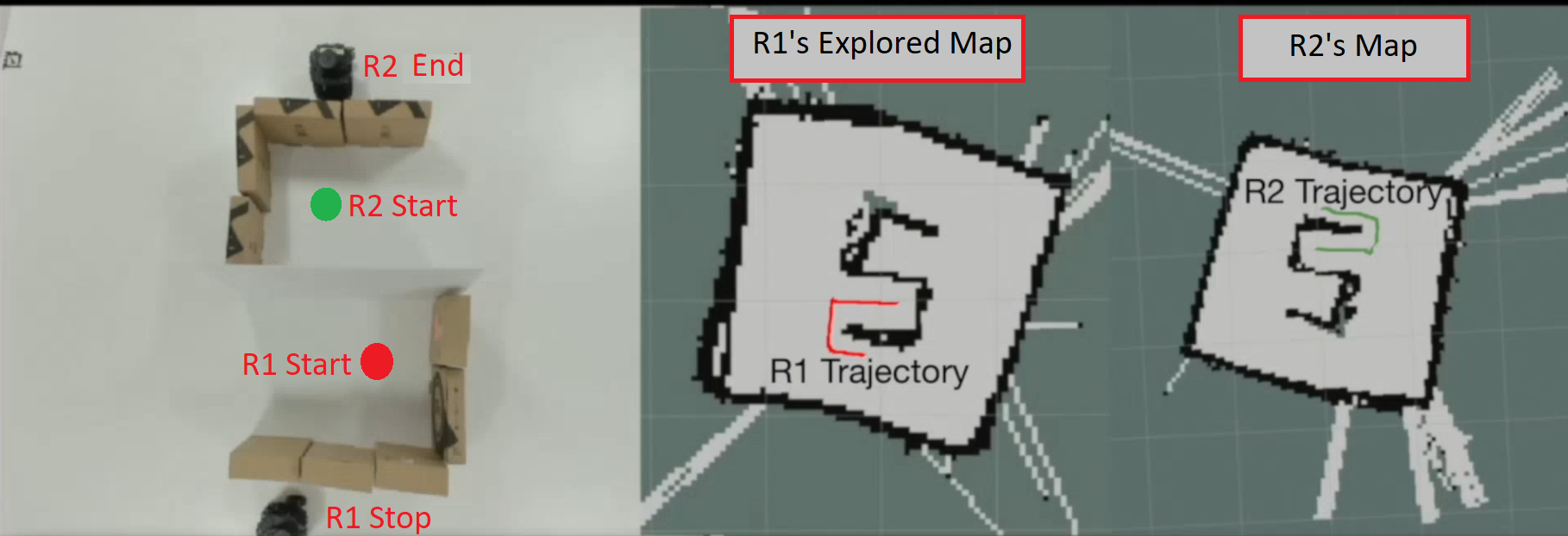}
    \caption{Sample outcome of the CQLite exploration with two Turtlebot3 robots in a real-world scenario with complex obstacle configuration.}
    \label{fig:hardware-sample}
\end{figure}

One of the limitations of the proposed CQLite is that it relies on wireless communication, which can be intermittent or harsh in specific real-world situations. In such scenarios, a communication-aware strategy can be integrated with our approach to tolerate changes in communication channels. 

\subsection{Real Robot Demonstration}
Two Turtlebot3 robots are used for real-world map exploration using the ROS Noetic framework. Robots can share information about the odometry and map output of their respective SLAM by subscribing to specific topic messages in ROS. Experiments are performed in a small-scale lab setting of $10m^2$. 
We tested the CQLite under two scenarios: 1) a simplistic setup where one robot can cover the entire map without the other robot needing to move; 2) a complex setup where obstacles obstruct both robots in their initial positions, necessitating both robots' contributions (see Fig.~\ref{fig:hardware-sample}). Both scenarios were successfully tested by exploring the entire map, as can be seen in this video: \url{https://youtu.be/n3unL1nuieQ}. We believe this provides further evidence for the practicality of the proposed exploration approach.

\section{Conclusion}
This paper proposed CQLite, a distributed Q-learning strategy for multi-robot exploration created to get around the excessive communication and computation complexity and expense of learning-based systems. CQLite, which reduces communication and processing overhead by simply sharing Q-value and mapping data over the network, performs well in practical applications. Experimental results revealed that it ensures comprehensive coverage, quick convergence, and cheaper computing costs compared to well-liked RRT and DRL techniques. The same mapping efficiency was attained with only half the CPU load and 80\% less communication overhead. In the future, we will examine the reward generality of CQLite to cope with various multi-robot applications.


\bibliography{ref}

\begin{thebibliography}{10}
\providecommand{\url}[1]{#1}
\csname url@samestyle\endcsname
\providecommand{\newblock}{\relax}
\providecommand{\bibinfo}[2]{#2}
\providecommand{\BIBentrySTDinterwordspacing}{\spaceskip=0pt\relax}
\providecommand{\BIBentryALTinterwordstretchfactor}{4}
\providecommand{\BIBentryALTinterwordspacing}{\spaceskip=\fontdimen2\font plus
\BIBentryALTinterwordstretchfactor\fontdimen3\font minus
  \fontdimen4\font\relax}
\providecommand{\BIBforeignlanguage}[2]{{%
\expandafter\ifx\csname l@#1\endcsname\relax
\typeout{** WARNING: IEEEtran.bst: No hyphenation pattern has been}%
\typeout{** loaded for the language `#1'. Using the pattern for}%
\typeout{** the default language instead.}%
\else
\language=\csname l@#1\endcsname
\fi
#2}}
\providecommand{\BIBdecl}{\relax}
\BIBdecl

\bibitem{burgard2005coordinated}
W.~Burgard, M.~Moors, C.~Stachniss, and F.~E. Schneider, ``Coordinated
  multi-robot exploration,'' \emph{IEEE Transactions on robotics}, vol.~21,
  no.~3, pp. 376--386, 2005.

\bibitem{parasuraman2017new}
R.~Parasuraman, S.~Caccamo, F.~B{\aa}berg, P.~{\"O}gren, and M.~Neerincx, ``A
  new ugv teleoperation interface for improved awareness of network
  connectivity and physical surroundings,'' \emph{Journal of Human-Robot
  Interaction}, vol.~6, no.~3, pp. 48--70, 2017.

\bibitem{fang2019autonomous}
B.~Fang, J.~Ding, and Z.~Wang, ``Autonomous robotic exploration based on
  frontier point optimization and multistep path planning,'' \emph{IEEE
  Access}, vol.~7, pp. 46\,104--46\,113, 2019.

\bibitem{dai2020fast}
A.~Dai, S.~Papatheodorou, N.~Funk, D.~Tzoumanikas, and S.~Leutenegger, ``Fast
  frontier-based information-driven autonomous exploration with an mav,'' in
  \emph{2020 IEEE international conference on robotics and automation
  (ICRA)}.\hskip 1em plus 0.5em minus 0.4em\relax IEEE, 2020, pp. 9570--9576.

\bibitem{latif2023communication}
E.~Latif, W.~Song, and R.~Parasuraman, ``Communication-efficient reinforcement
  learning in swarm robotic networks for maze exploration,'' in \emph{IEEE
  INFOCOM 2023 - IEEE Conference on Computer Communications Workshops (INFOCOM
  WKSHPS)}, 2023, pp. 1--6.

\bibitem{shrestha2019learned}
R.~Shrestha, F.-P. Tian, W.~Feng, P.~Tan, and R.~Vaughan, ``Learned map
  prediction for enhanced mobile robot exploration,'' in \emph{2019
  International Conference on Robotics and Automation (ICRA)}.\hskip 1em plus
  0.5em minus 0.4em\relax IEEE, 2019, pp. 1197--1204.

\bibitem{zhang2022multi}
Z.~Zhang, X.~Wang, Q.~Zhang, and T.~Hu, ``Multi-robot cooperative pursuit via
  potential field-enhanced reinforcement learning,'' in \emph{2022
  International Conference on Robotics and Automation (ICRA)}.\hskip 1em plus
  0.5em minus 0.4em\relax IEEE, 2022, pp. 8808--8814.

\bibitem{tolstaya2021multi}
E.~Tolstaya, J.~Paulos, V.~Kumar, and A.~Ribeiro, ``Multi-robot coverage and
  exploration using spatial graph neural networks,'' in \emph{2021 IEEE/RSJ
  International Conference on Intelligent Robots and Systems (IROS)}.\hskip 1em
  plus 0.5em minus 0.4em\relax IEEE, 2021, pp. 8944--8950.

\bibitem{yang2019self}
Q.~Yang, Z.~Luo, W.~Song, and R.~Parasuraman, ``Self-reactive planning of
  multi-robots with dynamic task assignments,'' in \emph{2019 International
  Symposium on Multi-Robot and Multi-Agent Systems (MRS)}.\hskip 1em plus 0.5em
  minus 0.4em\relax IEEE, 2019, pp. 89--91.

\bibitem{sadhu2018efficient}
A.~K. Sadhu and A.~Konar, ``An efficient computing of correlated equilibrium
  for cooperative $ q $-learning-based multi-robot planning,'' \emph{IEEE
  Transactions on Systems, Man, and Cybernetics: Systems}, vol.~50, no.~8, pp.
  2779--2794, 2018.

\bibitem{zhang2020rapidly}
L.~Zhang, Z.~Lin, J.~Wang, and B.~He, ``Rapidly-exploring random trees
  multi-robot map exploration under optimization framework,'' \emph{Robotics
  and Autonomous Systems}, vol. 131, p. 103565, 2020.

\bibitem{hu2020voronoi}
J.~Hu, H.~Niu, J.~Carrasco, B.~Lennox, and F.~Arvin, ``Voronoi-based
  multi-robot autonomous exploration in unknown environments via deep
  reinforcement learning,'' \emph{IEEE Transactions on Vehicular Technology},
  vol.~69, no.~12, pp. 14\,413--14\,423, 2020.

\bibitem{gao2018improved}
W.~Gao, M.~Booker, A.~Adiwahono, M.~Yuan, J.~Wang, and Y.~W. Yun, ``An improved
  frontier-based approach for autonomous exploration,'' in \emph{2018 15th
  International Conference on Control, Automation, Robotics and Vision
  (ICARCV)}.\hskip 1em plus 0.5em minus 0.4em\relax IEEE, 2018, pp. 292--297.

\bibitem{bouman2020autonomous}
A.~Bouman, M.~F. Ginting, N.~Alatur, M.~Palieri, D.~D. Fan, T.~Touma,
  T.~Pailevanian, S.-K. Kim, K.~Otsu, J.~Burdick \emph{et~al.}, ``Autonomous
  spot: Long-range autonomous exploration of extreme environments with legged
  locomotion,'' in \emph{IEEE/RSJ International Conference on Intelligent
  Robots and Systems (IROS)}.\hskip 1em plus 0.5em minus 0.4em\relax IEEE,
  2020, pp. 2518--2525.

\bibitem{latif2023seal}
E.~Latif and R.~Parasuraman, ``Seal: Simultaneous exploration and localization
  for multi-robot systems,'' in \emph{2023 IEEE/RSJ International Conference on
  Intelligent Robots and Systems (IROS)}, 2023, pp. 5358--5365.

\bibitem{zhang2022mr}
Z.~Zhang, J.~Yu, J.~Tang, Y.~Xu, and Y.~Wang, ``Mr-topomap: Multi-robot
  exploration based on topological map in communication restricted
  environment,'' \emph{IEEE Robotics and Automation Letters}, vol.~7, no.~4,
  pp. 10\,794--10\,801, 2022.

\bibitem{masaba2021gvgexp}
K.~Masaba and A.~Q. Li, ``Gvgexp: Communication-constrained multi-robot
  exploration system based on generalized voronoi graphs,'' in \emph{2021
  International Symposium on Multi-Robot and Multi-Agent Systems (MRS)}.\hskip
  1em plus 0.5em minus 0.4em\relax IEEE, 2021, pp. 146--154.

\bibitem{corah2019communication}
M.~Corah, C.~O’Meadhra, K.~Goel, and N.~Michael, ``Communication-efficient
  planning and mapping for multi-robot exploration in large environments,''
  \emph{IEEE Robotics and Automation Letters}, vol.~4, no.~2, pp. 1715--1721,
  2019.

\bibitem{gao2022meeting}
Y.~Gao, Y.~Wang, X.~Zhong, T.~Yang, M.~Wang, Z.~Xu, Y.~Wang, Y.~Lin, C.~Xu, and
  F.~Gao, ``Meeting-merging-mission: A multi-robot coordinate framework for
  large-scale communication-limited exploration,'' in \emph{2022 IEEE/RSJ
  International Conference on Intelligent Robots and Systems (IROS)}.\hskip 1em
  plus 0.5em minus 0.4em\relax IEEE, 2022, pp. 13\,700--13\,707.

\bibitem{serra2020whom}
A.~Serra-G{\'o}mez, B.~Brito, H.~Zhu, J.~J. Chung, and J.~Alonso-Mora, ``With
  whom to communicate: learning efficient communication for multi-robot
  collision avoidance,'' in \emph{2020 IEEE/RSJ International Conference on
  Intelligent Robots and Systems (IROS)}.\hskip 1em plus 0.5em minus
  0.4em\relax IEEE, 2020, pp. 11\,770--11\,776.

\bibitem{han2020cooperative}
R.~Han, S.~Chen, and Q.~Hao, ``Cooperative multi-robot navigation in dynamic
  environment with deep reinforcement learning,'' in \emph{2020 IEEE
  International Conference on Robotics and Automation (ICRA)}.\hskip 1em plus
  0.5em minus 0.4em\relax IEEE, 2020, pp. 448--454.

\bibitem{yu2019navigation}
X.~Yu, Y.~Wu, and X.-M. Sun, ``A navigation scheme for a random maze using
  reinforcement learning with quadrotor vision,'' in \emph{2019 18th European
  Control Conference (ECC)}.\hskip 1em plus 0.5em minus 0.4em\relax IEEE, 2019,
  pp. 518--523.

\bibitem{liu2022robust}
J.~Liu, K.~Chen, R.~Liu, Y.~Yang, Z.~Wang, and J.~Zhang, ``Robust and accurate
  multi-agent slam with efficient communication for smart mobiles,'' in
  \emph{2022 International Conference on Robotics and Automation (ICRA)}.\hskip
  1em plus 0.5em minus 0.4em\relax IEEE, 2022, pp. 2782--2788.

\bibitem{bernreiter2022collaborative}
L.~Bernreiter, S.~Khattak, L.~Ott, R.~Siegwart, M.~Hutter, and C.~Cadena,
  ``Collaborative robot mapping using spectral graph analysis,'' in \emph{2022
  International Conference on Robotics and Automation (ICRA)}.\hskip 1em plus
  0.5em minus 0.4em\relax IEEE, 2022, pp. 3662--3668.

\bibitem{zhao2021general}
M.~Zhao, X.~Guo, L.~Song, B.~Qin, X.~Shi, G.~H. Lee, and G.~Sun, ``A general
  framework for lifelong localization and mapping in changing environment,'' in
  \emph{2021 IEEE/RSJ International Conference on Intelligent Robots and
  Systems (IROS)}.\hskip 1em plus 0.5em minus 0.4em\relax IEEE, 2021, pp.
  3305--3312.

\bibitem{jia2021lvio}
Y.~Jia, H.~Luo, F.~Zhao, G.~Jiang, Y.~Li, J.~Yan, Z.~Jiang, and Z.~Wang,
  ``Lvio-fusion: A self-adaptive multi-sensor fusion slam framework using
  actor-critic method,'' in \emph{2021 IEEE/RSJ International Conference on
  Intelligent Robots and Systems (IROS)}.\hskip 1em plus 0.5em minus
  0.4em\relax IEEE, 2021, pp. 286--293.

\bibitem{mangelson2018pairwise}
J.~G. Mangelson, D.~Dominic, R.~M. Eustice, and R.~Vasudevan, ``Pairwise
  consistent measurement set maximization for robust multi-robot map merging,''
  in \emph{2018 IEEE international conference on robotics and automation
  (ICRA)}.\hskip 1em plus 0.5em minus 0.4em\relax IEEE, 2018, pp. 2916--2923.

\bibitem{keidar2014efficient}
M.~Keidar and G.~A. Kaminka, ``Efficient frontier detection for robot
  exploration,'' \emph{The International Journal of Robotics Research},
  vol.~33, no.~2, pp. 215--236, 2014.

\bibitem{uykan2019working}
Z.~Uykan, ``On the working principle of the hopfield neural networks and its
  equivalence to the gadia in optimization,'' \emph{IEEE Transactions on Neural
  Networks and Learning Systems}, vol.~31, no.~9, pp. 3294--3304, 2019.

\bibitem{li2020faster}
J.~Li, ``Faster parallel algorithm for approximate shortest path,'' in
  \emph{Proceedings of the 52nd Annual ACM SIGACT Symposium on Theory of
  Computing}, 2020, pp. 308--321.

\bibitem{ov2020impact}
S.~S. OV, R.~Parasuraman, and R.~Pidaparti, ``Impact of heterogeneity in
  multi-robot systems on collective behaviors studied using a search and rescue
  problem,'' in \emph{2020 IEEE International Symposium on Safety, Security,
  and Rescue Robotics (SSRR)}.\hskip 1em plus 0.5em minus 0.4em\relax IEEE,
  2020, pp. 290--297.

\bibitem{duda2014novel}
P.~Duda, M.~Jaworski, L.~Pietruczuk, and L.~Rutkowski, ``A novel application of
  hoeffding's inequality to decision trees construction for data streams,'' in
  \emph{2014 International Joint Conference on Neural Networks (IJCNN)}.\hskip
  1em plus 0.5em minus 0.4em\relax IEEE, 2014, pp. 3324--3330.

\bibitem{konda2020decentralized}
R.~Konda, H.~M. La, and J.~Zhang, ``Decentralized function approximated
  q-learning in multi-robot systems for predator avoidance,'' \emph{IEEE
  Robotics and Automation Letters}, vol.~5, no.~4, pp. 6342--6349, 2020.

\end{thebibliography}
\bibliographystyle{IEEEtran}

\end{document}